\documentclass[lettersize,journal]{IEEEtran}
\usepackage{amsmath,amsfonts}
\usepackage{algorithmic}
\usepackage{array}
\usepackage[caption=false,font=normalsize,labelfont=sf,textfont=sf]{subfig}
\usepackage{textcomp}
\usepackage{stfloats}
\usepackage{url}
\usepackage{verbatim}
\usepackage{graphicx}
\usepackage{cite}
\usepackage{enumitem}
\usepackage{multirow}
\usepackage{multicol}
\usepackage{amsmath,amssymb,mathrsfs}
\usepackage{booktabs}

\usepackage[sort,numbers]{natbib}
\usepackage{graphicx}
\usepackage[colorlinks,
            linkcolor=black,
            anchorcolor=black,
            citecolor=black]{hyperref}
\usepackage{amsmath}
\usepackage{color}
\usepackage{graphicx} 
\usepackage{float} 
\usepackage{subcaption} 
\usepackage{comment}
\usepackage[noline, ruled, vlined]{algorithm2e}
\usepackage{multirow}
\usepackage{booktabs}
\usepackage{tabularx}

\usepackage{amsmath} 
\usepackage{amssymb} 

\usepackage{enumitem}
\usepackage{multirow}
\usepackage{multicol}
\usepackage{amsmath,amssymb,mathrsfs}
\usepackage{booktabs}

\usepackage{
amsfonts,
cite,
pifont,
graphicx,
multirow,
booktabs,
amsmath,
amssymb,
hhline,
url,
hyperref}
\usepackage[table,xcdraw]{xcolor}

\newcommand\shline{\specialrule{0.8pt}{0pt}{0pt}}

\usepackage{xcolor}
\definecolor{darkblue}{rgb}{0, 0, 0.6}
\definecolor{darkgreen}{rgb}{0, 0.7, 0}
\definecolor{darkred}{rgb}{0.8, 0, 0}

\let\oldcite\cite
\renewcommand{\cite}[1]{\textcolor{darkblue}{\oldcite{#1}}}

\newcommand{\best}[1]{\textbf{\textcolor{red}{#1}}}
\newcommand{\secondbest}[1]{\underline{\textcolor{blue}{#1}}}

\newcommand*\colorcheck{%
  \expandafter\newcommand\csname greencheck\endcsname{\textcolor{darkgreen}{\ding{52}}}%
}
\newcommand*\colorcross{%
  \expandafter\newcommand\csname redcross\endcsname{\textcolor{darkred}{\ding{56}}}%
}
\colorcheck
\colorcross

\ifCLASSINFOpdf
\else
\fi
\begin{document}
\bibliographystyle{plain}
%
\title{Capturing Stable HDR Videos Using a Dual-Camera System}
%
%
%

\author{Qianyu Zhang, Bolun Zheng*,~\IEEEmembership{Member, IEEE}, Lingyu Zhu, Hangjia Pan, Zunjie Zhu, \\ 
Zongpeng Li,~\IEEEmembership{Senior Member, IEEE}, 
Shiqi Wang,~\IEEEmembership{Senior Member, IEEE}

\thanks{Qianyu Zhang, Bolun Zheng, Hangjia Pan, Zongpeng Li are with the School of Automation, Hangzhou Dianzi University, Hangzhou 310018, China (e-mail: qyzhang@hdu.edu.cn; blzheng@hdu.edu.cn; 1072272964@qq.com; zongpeng@tsinghua.edu.cn).
\\
Zunjie Zhu is with the Lishui Institute of Hangzhou Dianzi University (e-mail: zunjiezhu@hdu.edu.cn).
\\
Lingyu Zhu and Shiqi Wang are with the Department of Computer Science, City University of Hong Kong (e-mail:lingyzhu-c@my.cityu.edu.hk; shiqwang@cityu.edu.hk)
}
}

%
%

\markboth{Journal of \LaTeX\ Class Files,~Vol.~14, No.~8, August~2015}%
{Shell \MakeLowercase{\textit{et al.}}: Bare Demo of IEEEtran.cls for IEEE Journals}
%



\maketitle

\begin{abstract}
High Dynamic Range (HDR) video acquisition using the alternating exposure (AE) paradigm has garnered significant attention due to its cost-effectiveness with a single consumer camera. However, despite progress driven by deep neural networks, these methods remain prone to temporal flicker in real-world applications due to inter-frame exposure inconsistencies. To address this challenge while maintaining the cost-effectiveness of the AE paradigm, we propose a novel learning-based HDR video generation solution. Specifically, we propose a dual-stream HDR video generation paradigm that decouples temporal luminance anchoring from exposure-variant detail reconstruction, overcoming the inherent limitations of the AE paradigm. To support this, we design an asynchronous dual-camera system (DCS), which enables independent exposure control across two cameras, eliminating the need for synchronization typically required in traditional multi-camera setups. Furthermore, an exposure-adaptive fusion network (EAFNet) is formulated for the DCS system. EAFNet integrates a pre-alignment subnetwork that aligns features across varying exposures, ensuring robust feature extraction for subsequent fusion, an asymmetric cross-feature fusion subnetwork that emphasizes reference-based attention to effectively merge these features across exposures, and a restoration subnetwork for final output. Extensive experimental evaluations demonstrate that the proposed method achieves state-of-the-art performance across various datasets, showing the remarkable potential of our solution in HDR video reconstruction. The codes and data captured by DCS will be available at \url{https://zqqqyu.github.io/DCS-HDR/}.
\end{abstract}


\IEEEpeerreviewmaketitle

\section{Introduction}
Capturing the full range of illumination in high dynamic range (HDR) scenarios is a challenging task for standard digital cameras~\cite{debevec2008recovering, jacobs}.
Thanks to the multi-exposure fusion (MEF) technology \cite{bogoni2000extending, sen2012robust}, we can successfully capture HDR images in static scenarios by fusing multiple images of different exposure times.
However, capturing HDR videos with a standard digital camera in dynamic scenes has not yet been effectively resolved \cite{wang2021deep, wang2024pix2hdr}.
Early studies concentrated on designing hardware systems for directly capturing HDR video, including internal/external beam splitters~\cite{ mcguire2007optical, tocci2011versatile}, specialized sensors \cite{choi2017reconstructing, heide2014flexisp, martel2020neural}, modular cameras \cite{zhao2015unbounded} and neuromorphic cameras \cite{han2020neuromorphic}, \textit{etc.}
Although these solutions show good potential for capturing high-quality HDR videos, their high cost and bulky design severely limit their widespread adoption.

\begin{figure*}[t] 
\centering 
\includegraphics[width=1\textwidth]{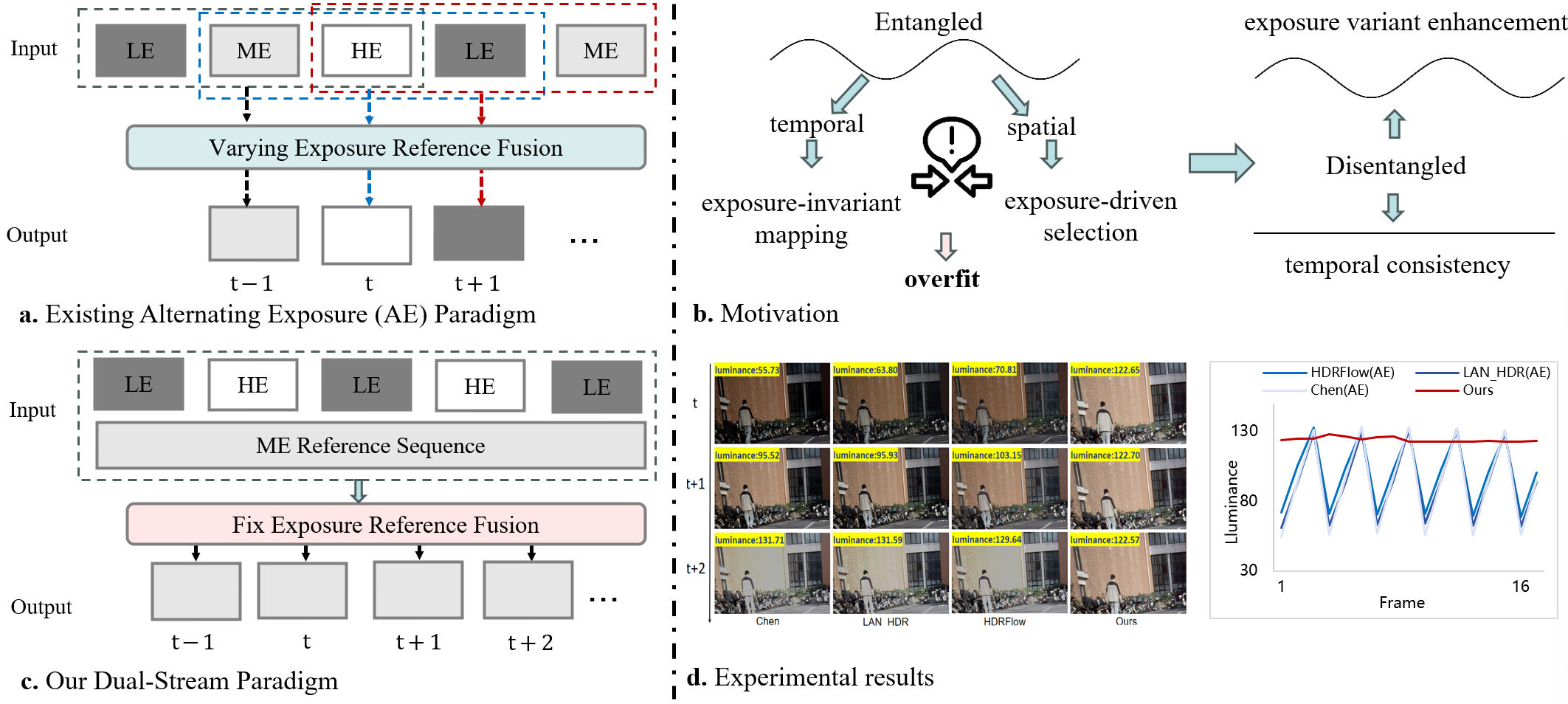} 
\caption{
Overview of the alternating exposure (AE) paradigm and our proposed dual-stream paradigm.
(a) existing AE paradigm fuses alternating exposures (LE/ME/HE) with varying reference frames, resulting in flickering artifacts (luminance fluctuations exceeding 30).
(b-d) Our dual-stream paradigm employs a fixed-exposure reference sequence for stable luminance anchoring, while a separate stream handles exposure-variant enhancement, ensuring consistent illumination across reference frames, enabling temporally consistent HDR reconstruction results (luminance fluctuations limited to under 1).} 
\label{intro} 
\end{figure*}

The alternating exposure (AE) paradigm, first introduced by Kang \textit{et al.} \cite{kang2003high}, enables low-cost HDR video acquisition using a single camera with rapidly switching exposures, but it also introduces challenges in maintaining temporal consistency and suppressing ghosting artifacts. 
Traditional AE-based methods adopt strategies such as motion-compensated fusion~\cite{kalantari2013patch, gryaditskaya2015motion} and optical flow-based reconstruction~\cite{mangiat2010high} to align and merge complementary exposures for HDR reconstruction, but they remain highly sensitive to motion misalignment, struggle with occlusions and non-rigid deformations, and often rely on hand-crafted heuristics that lack generalization. Recent deep learning approaches \cite{kalantari2019deep, cui2024exposure, xu2024hdrflow} have advanced this direction by employing deep neural architectures with stronger feature representations and learning-based alignment mechanisms. 
These methods have achieved promising results by training on synthetic datasets with well-structured exposure patterns and often produce temporally consistent outputs under controlled settings. However, in real-world scenarios, variations in brightness distribution caused by changes in scene content, illumination, or sensor response can disrupt the learned alignment and fusion behavior. As a result, networks that perform well in ideal conditions often suffer from temporal flickering or inconsistent reconstruction when deployed in real environments (see Fig.~\ref{intro}(d)).

This instability highlights a fundamental limitation of the AE paradigm itself, rather than just model design, thus leading us to explore a new solution that circumvents inter-frame exposure inconsistency.
Specifically, in the AE paradigm, detail selection and global luminance determination are inherently entangled at the input level.
Existing methods \cite{cui2024exposure, xu2024hdrflow} typically process both aspects within a single modeling stream, without explicit disentanglement in input design or feature representation.
As a result, the network must simultaneously decide which regions to source from each exposure and how to establish the overall luminance mapping within the same representational space.
As shown in Fig.~\ref{intro}, this tight coupling leads to feature interference and makes the learned mapping strongly dependent on the exposure patterns seen during training, thereby reducing robustness and temporal stability under distribution shifts.
To address this, we propose a dual-stream generation paradigm that decouples temporal luminance anchoring from exposure-variant detail reconstruction, thereby reducing feature interference and enhancing generalization across varying exposure patterns and scene conditions.
Our paradigm is meticulously designed at both the hardware system and algorithmic levels, forming a complete HDR video generation solution that supports practical deployment and robust reconstruction in real-world environments.

Our hardware architecture is designed to implement an asynchronous dual-camera system (DCS) that capitalizes on the prevalent availability of multi-camera hardware in consumer electronic devices. Unlike prior dual-camera HDR imaging systems \cite{dong2021miehdr, li2023dual} designed for static scenes under strict synchronization constraints, our DCS is tailored for dynamic video scenarios. By relaxing the synchronization requirement, it supports independent exposure control across two cameras. 
This design not only facilitates flexible and practical deployment without precision alignment but also avoids the frame rate bottleneck imposed by long-exposure frames in synchronized systems. As a result, our system enables high-quality HDR video capture at high frame rates.
Specifically, the designed DCS system utilizes a fixed-exposure reference sequence as the temporal baseline, while simultaneously leveraging a varying-exposure stream to enrich the dynamic range.
This system-level approach addresses key limitations of existing video methods. 
The reference stream mitigates the temporal inconsistencies and flickering commonly observed in AE-based single-camera systems, 
while the division of labor between the two cameras eliminates the need for precise synchronization inherent in traditional multi-camera systems, allowing for independent and flexible exposure control. 
Furthermore, the proposed DCS remains fully compatible with most existing image HDR deghosting algorithms, enabling temporally consistent and robust HDR video reconstruction.

Building on the proposed paradigm and DCS system, we now turn to the algorithmic level. We first revisit existing HDR deghosting approaches to extract insights and expose their limitations.
Current methods can be broadly divided into optical-flow-based \cite{prabhakar2019fast,peng2018deep} and attention-based \cite{yan2019attention,liu2022ghost} strategies. 
The former relies on optical flow to align inputs before fusion, which often suffers from low accuracy in estimating optical flow.  
On the other hand, attention-based methods, while offering greater flexibility, their attention maps are typically computed only from texture and luminance cues, without explicitly leveraging exposure information. 
Moreover, they generally assign equal importance to reference and non-reference frames, which allows attention to be skewed by non-reference inputs in complex scenes. 
In our DCS, the presence of parallax introduces additional variations between inputs, making it even more important to design attention mechanisms that emphasize reference frames.
Consequently, both categories exhibit limited robustness in dynamic, exposure-varying, and parallax-affected settings, where inaccurate alignment and exposure-insensitive attention lead to ghosting and color distortion.
To address this, we propose an exposure-adaptive fusion network (EAFNet) tailored for the proposed DCS system to achieve robust and generalized HDR reconstruction. 
Our EAFNet consists of a pre-alignment subnetwork, an asymmetric cross-feature fusion subnetwork, and a restoration subnetwork. 
The pre-alignment subnetwork consists of a global luminance alignment (GLA) and an exposure-guided feature selection module (EFSM) to leverage exposure information, enhancing meaningful details and ensuring that the most relevant features are preserved despite varying exposure levels.
The asymmetric cross-feature fusion subnetwork aims to explore reference-dominated attention maps, align cross-scale features, and integrate cross-feature information to improve the fusion. 
Additionally, the refined features are reconstructed by a lightweight restoration subnetwork built upon simplified blocks from existing architectures.
In summary, our contributions can be summarized as follows:
\begin{itemize}
    \item 
   \textbf{Paradigm}: We introduce a dual-stream HDR video generation paradigm that explicitly decouples temporal luminance anchoring from exposure-variant detail reconstruction. Our approach employs a fixed-exposure stream to maintain temporal alignment across frames, while a complementary stream with varying exposures enhances the dynamic range. This design fundamentally improves temporal consistency and reconstruction stability.

    \item
    \textbf{System}: We design a dual-camera system to validate the feasibility of our proposed solution and bridge the gap between algorithmic design and practical deployment. Unlike traditional synchronized setups constrained by long-exposure frames, our system enables high-frame-rate video capture in dynamic scenes by supporting independent exposure control without requiring hardware-level synchronization. Moreover, the system seamlessly integrates with existing image deghosting methods to achieve temporally consistent video reconstruction.

    \item 
    \textbf{Method}: To support our dual-camera system, we propose a novel model design, EAFNet. We leverage exposure information and explore the intrinsic properties of the images, ensuring that the most relevant features receive attention under different exposure conditions. Furthermore, we explore reference-dominated attention and effectively fuse complementary features across exposures, enabling robust representation under diverse lighting scenarios.

    \item 
    \textbf{Results}: Our proposed dual-stream HDR video generation paradigm demonstrates significant advancements across multiple aspects. Through extensive experimental results, we demonstrate its effectiveness. 

\end{itemize}

The rest of this article is organized as follows. 
Section~\ref{RELATED}
presents the related works. 
Section~\ref{system} introduces our dual-camera system for video HDR capturing.
Section~\ref{method} introduces the proposed
EAFNet in detail.
Section~\ref{experiment} presents the experimental results, and the conclusion is given in Section~\ref{conclusion}.

\section{RELATED WORK}
\label{RELATED}

\subsection{HDR Image Reconstruction.}

\subsubsection{Traditional Approaches}

Traditional multi-exposure HDR image fusion approaches address the ghosting problem using two main approaches: motion-aligned-based and motion-rejection-based methods. Motion-aligned methods focus on globally aligning LDR images, followed by rejection of unaligned pixels to eliminate ghosting. Early works by Bogoni \textit{et al.} \cite{bogoni2000extending} used optical flow for motion estimation, while later improvements by Kang \textit{et al.} \cite{kang2003high} incorporated exposure time information for enhanced alignment. Further advancements include patch-based energy minimization by Sen \textit{et al.} \cite{sen2012robust} and luminance-based optimization by Hu \textit{et al.} \cite{hu2013hdr}. Motion-rejection methods, on the other hand, identify and exclude motion regions before fusion. Grosch \textit{et al.} \cite{grosch2006fast} and Jacobs \textit{et al.} \cite{jacobs2008automatic} detected motion using intensity differences and weighted variance, respectively, while He \textit{et al.} \cite{heo2010ghost} and Zhang \textit{et al.} \cite{zhang2011gradient} refined motion detection with graph-cut and image gradient techniques. 

\subsubsection{Deep Learning Approaches}
These approaches typically evolve from rigid motion compensation to flexible feature fusion, and recently to implicit representations.

\noindent \textbf{CNN-based Alignment and Reconstruction.} 
Early approaches focused on explicitly handling motion between frames. Kalantari \textit{et al.} \cite{kalantari1} pioneered the use of optical flow to register non-reference LDR images. Recognizing that traditional flow fails in saturated regions, subsequent works \cite{peng2018deep, prabhakar2019fast} and SAFNet \cite{kong2024safnet} employed deep optical flow networks to refine motion estimation. Alternatively, flow-free methods treat HDR as a direct image transformation task, employing direct feature concatenation \cite{wu2018end, yan2021towards} or GAN-based frameworks \cite{niu2021hdr, li2022uphdr} to hallucinate missing details. Notably, Mattur \textit{et al.} \cite{mattur2021deep} extended this to stereo pairs for reconstruction. However, these methods often struggle with large-distance motions or severe occlusions, leading to ghosting artifacts when explicit alignment fails.

\noindent \textbf{Attention Mechanisms and Emerging Representations.} 
To overcome the limitations of rigid alignment, attention mechanisms were introduced to implicitly select and fuse beneficial features. AHDRNet \cite{yan2019attention} first utilized attention to suppress misaligned regions, while Liu \textit{et al.} \cite{liu2021adnet} introduced deformable convolutions for multi-scale alignment. Recently, Transformer-based architectures \cite{liu2022ghost, chen2023improving, yan2023unified, tel2023alignment} have integrated global context modeling to enhance feature representation. 
Beyond standard architectures, recent trends have also explored implicit neural representations \cite{wu2024fast, choi2025dual} to model continuous radiance fields for dynamic scenes. 
Despite these advances, most existing attention methods treat reference and non-reference features equally or lack explicit exposure modeling, which often results in suboptimal fusion in high-contrast regions dominated by reference-frame content.

\subsection{HDR Video Reconstruction} 
\noindent \textbf{Dedicated Hardware Solutions}. Dedicated hardware solutions, such as single-pixel imaging \cite{nayar2000camera}, scanline exposure/ISO \cite{choi2017reconstructing}, internal \cite{tocci2011versatile} or external beam splitters \cite{mcguire2007optical}, as well as residue cameras \cite{zhao2015unbounded} and neuromorphic vision cameras \cite{han2020neuromorphic}, are capable of rapidly and efficiently generating detailed HDR images or videos on specialized devices. However, the complexity and high cost of these hardware solutions limit their widespread adoption. 

\noindent \textbf{Traditional Methods}. 
Before the deep learning era, reconstruction primarily relied on aligning alternating exposure sequences. Kang \textit{et al.} \cite{kang2003high} pioneered this with global and local registration techniques. Subsequent works refined this via block-based motion estimation \cite{mangiat2010high, mangiat2011spatially} or adaptive weighting schemes \cite{gryaditskaya2015motion} to mitigate artifacts. Li \textit{et al.} \cite{li2016maximum} further proposed statistical methods to bypass exact alignment. However, these traditional approaches are computationally expensive and prone to visible ghosting artifacts in complex dynamic scenes.

\noindent \textbf{Deep Learning Methods.} 
Recent advancements have shifted focus to data-driven approaches. Kalantari \textit{et al.} \cite{kalantari2019deep} proposed the first CNN-based framework for alternating exposure HDR video. Following this, researchers have explored various alignment strategies:  
\textit{Flow and Attention-based Alignment:} HDRFlow \cite{xu2024hdrflow} introduced multi-scale large kernels for efficient flow estimation to handle large motions. To address flow inaccuracies, Chen \textit{et al.} \cite{chen2021hdr} integrated deformable convolutions for coarse-to-fine alignment, while LAN-HDR \cite{chung2023lan} utilized luminance-based attention to align adjacent frames with the reference. 
\textit{Alternative Strategies:} Some methods attempt to bypass direct cross-exposure alignment. Khan \textit{et al.} \cite{khan2022deephs} leveraged video frame interpolation to generate intermediate frames from same-exposure inputs, thereby avoiding complex motion estimation between different exposures. Similarly, Cui \textit{et al.} \cite{cui2024exposure} treated the problem as exposure completion, employing feature interpolation to render HDR content.

\textit{Limitation of Prior Arts:} Despite these improvements, a fundamental limitation remains: alternating exposure setups inherently introduce brightness inconsistencies in the reference stream (switching between short and long exposures). This leads to temporal instability and flickering. In contrast, our design adopts a fixed-exposure reference strategy, which naturally anchors brightness consistency and effectively mitigates flickering artifacts.

    \begin{figure}[!t] 
    \centering 
    \includegraphics[width=0.5\textwidth]{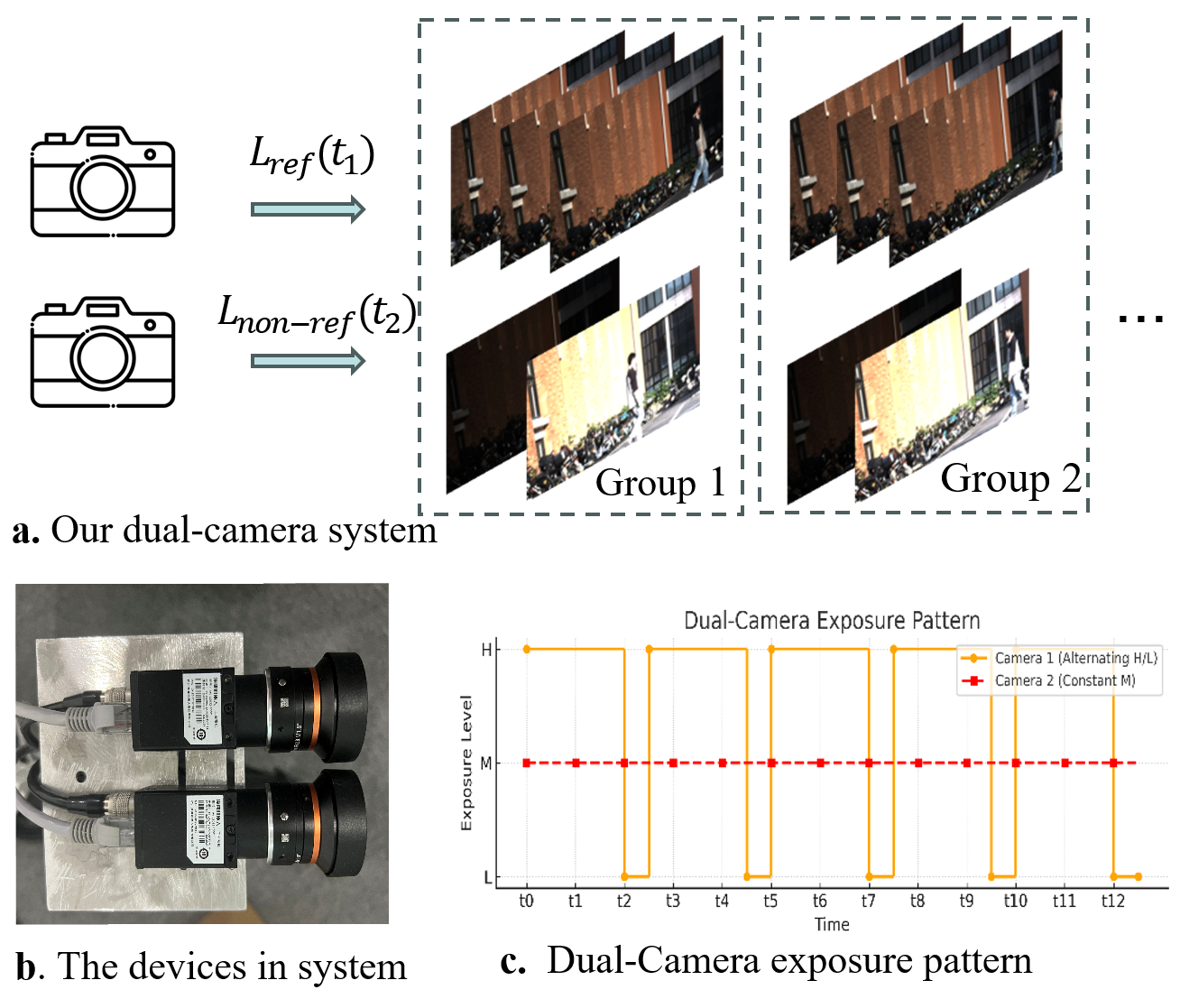} 
    \caption{Visualization of our dual-camera system. The primary camera captures continuous medium-exposure sequences as reference for temporal consistency, while the secondary camera alternates between low- and high-exposure to provide complementary information for reconstruction.} 
    \label{dural_result} 
    \end{figure}

\subsection{Dual-Camera Systems}

\noindent \textbf{Hardware-Centric and Event-Based Approaches.}
Early works \cite{mcnamee2015live} explored dual-system architectures for real-time HDR, while recent studies \cite{rebecq2019high, bao2024temporal} have leveraged event cameras for their superior temporal resolution. However, these hardware-centric approaches often rely on non-standard sensors or complex optical setups, limiting their widespread adoption. Consequently, the field has shifted towards deep learning-based methods using standard consumer cameras.

\noindent \textbf{Dual-Camera HDR.}
Deep learning-based dual-camera HDR methods \cite{dong2021miehdr, li2023dual} typically leverage stereo pairs to synthesize high-quality content. However, these approaches face two fundamental limitations. 
First, they operate under the strict assumption of 1D parallax on rectified stereo pairs. By constraining alignment to horizontal disparities, they limit their capacity to model complex, non-rigid distortions. Specifically, motion blur in long-exposure frames creates structural inconsistencies that cannot be resolved by disparity-based alignment alone. 
Second, they require strict temporal synchronization. This is challenging in dynamic scenes due to exposure differences, often leading to ghosting artifacts. Furthermore, the frame rate in synchronized systems is bounded by the longest exposure time, making them unsuitable for high-frame-rate capture.

\noindent \textbf{Dual-Camera Video Enhancement.}
Beyond HDR, dual-camera frameworks have been explored in related fields \cite{cheng2020dual, cheng2022h, dong2025video}. These methods typically pair a high-resolution/low-frame-rate stream with a low-resolution/high-frame-rate stream to reconstruct high spatiotemporal quality. While they exploit asymmetric spatial information, their optimization targets differ fundamentally from ours: they generally assume consistent exposure between cameras to maximize fidelity, rather than expanding dynamic range.

\section{DUAL-CAMERA SYSTEM FOR CAPTURING HDR VIDEO}
\label{system}

\noindent \textbf{Overview.} An inexpensive mainstream solution for HDR video reconstruction is generating video from image sequences captured at alternating exposure levels \cite{kalantari2019deep, cui2024exposure, xu2024hdrflow}. However, this approach inherits the exposure limitations of the reference frame, often resulting in visible artifacts and temporal flickering in the generated video, significantly constraining its practical application. To address this, we propose a dual-stream paradigm supported by a dual-camera system (as shown in Fig.~\ref{dural_result}), where the primary camera continuously captures medium-exposure frames to ensure temporal consistency, while the secondary camera captures low-exposure and high-exposure frames to provide complementary information.

\subsection{Dual-Camera System}
\noindent \textbf{System Configuration.} Our system employs two MV-CS032-10GC industrial cameras, which provide low-level access to operational controls for precise adjustment of exposure time and shooting strategies. Both cameras are equipped with identical lenses and are mounted side by side, secured by a mechanical support structure to ensure their relative positions remain stable throughout the imaging process. 

\noindent \textbf{Dual-Camera Calibration.} To eliminate distortion and ensure fusion, we perform camera calibration using a checkerboard pattern. Intrinsic and extrinsic parameters are computed from images captured by the dual cameras using the camera calibration toolbox \cite{zhang2002flexible}. The calibration parameters are then applied to register the images, resulting in the final aligned images. This reduces potential errors and improves the subsequent processing steps.

\noindent \textbf{Video Capture and System Design.} 
Our dual-camera system employs two identical cameras with resolution $w\times h$. In practical operation, the cameras are run asynchronously: one continuously records a medium-exposure reference sequence $L_{ref}(t_1)$ to provide temporally consistent baseline frames, while the other alternates between low and high exposures $L_{non\mbox{-}ref}(t_2)$ to supply the luminance diversity required for dynamic range expansion. To leverage the information from two sequences, we construct approximate pairings using timestamp metadata: as shown in Fig.~\ref{dural_result}(a), for each low/high exposure pair in the alternating stream, we gather reference frames falling within its defined temporal neighborhood and combine each of them with the same low/high pair to form input groups, enabling the reuse of one low/high pair across multiple reference frames. This pairing does not require frame-level precision, as the scenes are inherently dynamic and minor temporal offsets have a negligible impact on multi-exposure fusion. Each input group is then fed into the network to reconstruct HDR video at the same frame rate as $L_{ref}(t_1)$. 

Overall, the design formulates the task as multi-exposure fusion for dynamic scenes, leveraging the consistent exposure of the reference sequence to eliminate the need for luminance alignment between neighboring frames.
Existing HDR image deghosting methods can be seamlessly extended to achieve consistent luminance results. To address challenges from video data and the variability of real-world scenes, we propose EAFNet, a more robust network, detailed in Section \ref{method}.

\begin{figure*}[!t] 
\centering 
\includegraphics[width=0.95\textwidth]{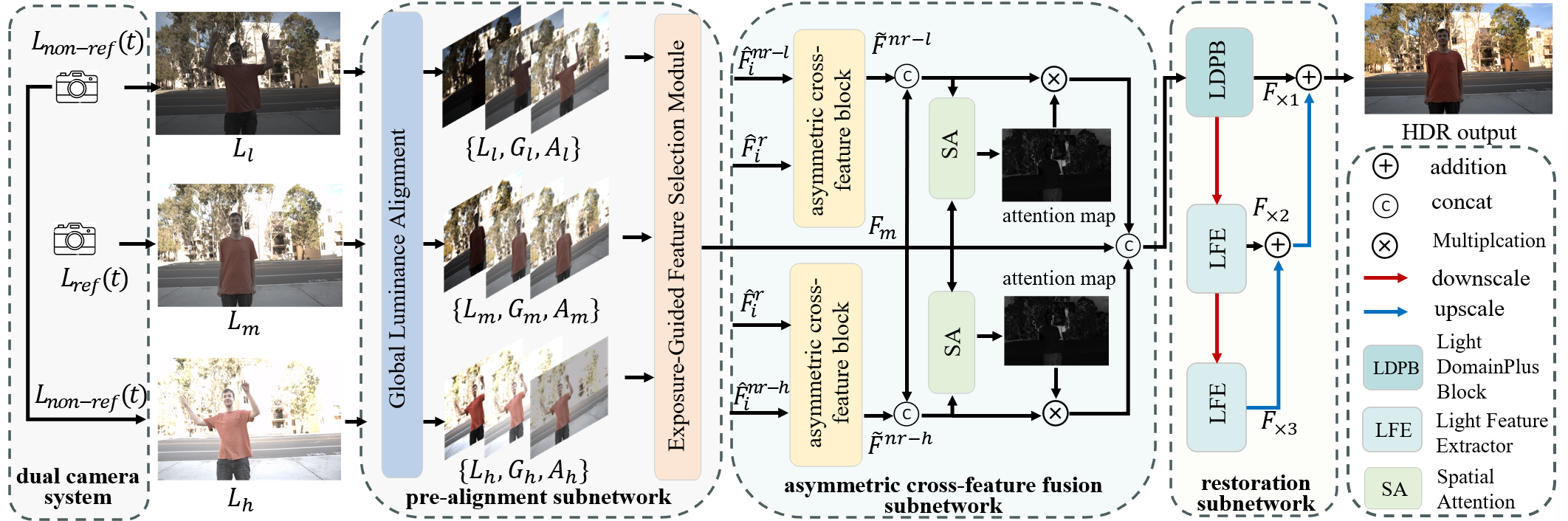} 
\caption{The architecture of EAFNet consists of a pre-alignment subnetwork, an asymmetric cross-feature fusion subnetwork, and a restoration subnetwork. We introduce GLA and EFSM to leverage exposure information, explore the intrinsic properties of the images, and help preserve finer details across varying exposures. The asymmetric cross-feature fusion subnetwork improves image fusion by aligning cross-scale features and performing cross-feature fusion. The restoration subnetwork adopts a multi-scale architecture to reduce ghosting and refine features at different resolutions.} 
\label{Fig2} 
\end{figure*}

\subsection{Further Discussion}

The proposed paradigm offers several distinct advantages.

First, the dual-stream design introduces an exposure-asymmetric configuration that explicitly decouples temporal stability from exposure diversity, thereby alleviating the feature interference inherent in AE-based HDR systems. Building upon this decoupling, the paradigm achieves strong scalability and generalization, allowing multi-exposure fusion models trained on public datasets to be directly applied without device-specific retraining, enabling plug-and-play deployment.


Second, it offers operational flexibility through its inherent tolerance to hardware asynchrony. Unlike synchronized systems that require precise trigger signals, our robust fusion mechanism enables the auxiliary stream to provide complementary dynamic range information regardless of its temporal phase relative to the reference stream. This relaxes strict sensor timing constraints and simplifies hardware deployment. Furthermore, by anchoring the output frame rate to the reference stream, the system can deliver high-frame-rate HDR video while the auxiliary stream operates at a lower rate with preset exposures. Such a design improves energy efficiency and reduces bandwidth and storage overhead, enabling seamless integration with standard video encoding protocols.



Finally, our paradigm extends naturally to mobile devices featuring homogeneous dual-camera arrays, potentially offering superior efficacy compared to our prototype.
On the one hand, the compact baseline of mobile cameras (approx. $1\sim2$ cm) is significantly smaller than that of our prototype (40 mm), naturally mitigating the parallax challenges analyzed in Sec.~\ref{Parallax Analysis}. 
On the other hand, our asynchronous design circumvents the strict synchronization constraints often hindered by mobile OS latencies. 
Crucially, this paradigm allows mobile ISPs to leverage our dual-stream protocol to feed existing alignment methods.

\section{METHOD}
\label{method}
\subsection{Overview}
In this section, we introduce the overall framework of the proposed EAFNet, built upon a pre-alignment subnetwork (in Sec.~\ref{pre-alignment subnetwork}), an asymmetric cross-feature fusion subnetwork (in Sec.~\ref{Asymmetric Cross-feature Fusion subnetwork}) and a restoration subnetwork (in Sec.~\ref{restoration subnetwork}).

As shown in Fig.~\ref{Fig2}, the pre-alignment subnetwork aims to improve detail preservation and color accuracy by exploring exposure information and extracting meaningful features from the non-reference image. It incorporates Global Luminance Alignment (GLA) to ensure consistent brightness across varying exposures, and an Exposure-guided Feature Selection Module (EFSM) to identify and focus on the most relevant regions based on the exposure information. These mechanisms work together to enhance the input features before fusion.
Next, the asymmetric cross-feature fusion subnetwork plays a critical role in blending features from different exposures while emphasizing the most important details. It processes the low- and high-exposure inputs through separate asymmetric cross-feature fusion blocks, each built with a multi-scale design. The fusion is driven by an Asymmetric Cross-Attention (ACA), which leverages reference-dominated attention maps to improve the alignment of query and key features, facilitating more effective feature propagation across multiple scales.
Finally, the restoration subnetwork addresses common issues such as ghosting artifacts. It is designed to suppress ghosting while preserving fine details at multiple scales.

\subsection{pre-alignment subnetwork}
\label{pre-alignment subnetwork}
The pre-alignment subnetwork aims to extract meaningful details from the non-reference image by leveraging exposure information. GLA is introduced to maintain consistent luminance across varying exposures, while the EFSM explores the correlation between inputs and exposure to focus on the most relevant regions for fusion.

\noindent \textbf{Global Luminance Alignment (GLA).}
Following established HDR reconstruction pipelines \cite{kalantari2017deep, yan2019attention}, we employ three LDR images $\{{L}_l, {L}_m, {L}_h\}$ (sorted by their exposure length) as input.
These images are first mapped to the HDR domain relying on gamma correction \cite{kalantari2017deep} to produce a set of gamma-corrected images ${{G}_i}$, where $i \in \{l, m, h\}$ and $\gamma = 2.2$. 

To reduce inter-exposure luminance discrepancies and improve alignment robustness, we introduce GLA in the pre-alignment stage. GLA operates in the sRGB domain and ensures consistent luminance distributions across inputs and minimizes exposure-induced mismatches during fusion.
Formally, for each non-reference LDR image \(L_i\), global alignment is performed as:
\begin{equation}
A_i = \operatorname{clip}\!\left( L_i \times \frac{\mathbb{E}(L_m)}{\mathbb{E}(L_i)}, \; 0, 1 \right), \quad i \in \{l, h\},
\label{eq:luminance_alignment}
\end{equation}
where \(A_i\) denotes the luminance-aligned image, \(\mathbb{E}(\cdot)\) computes the average luminance, and \(\operatorname{clip}(\cdot, 0, 1)\) restricts pixel values to the valid dynamic range. Since the mid-exposure image \(L_m\) serves as the reference, we set \(A_m = L_m\).

\begin{figure}[!t] 
    \centering 
\includegraphics[width=0.48\textwidth]{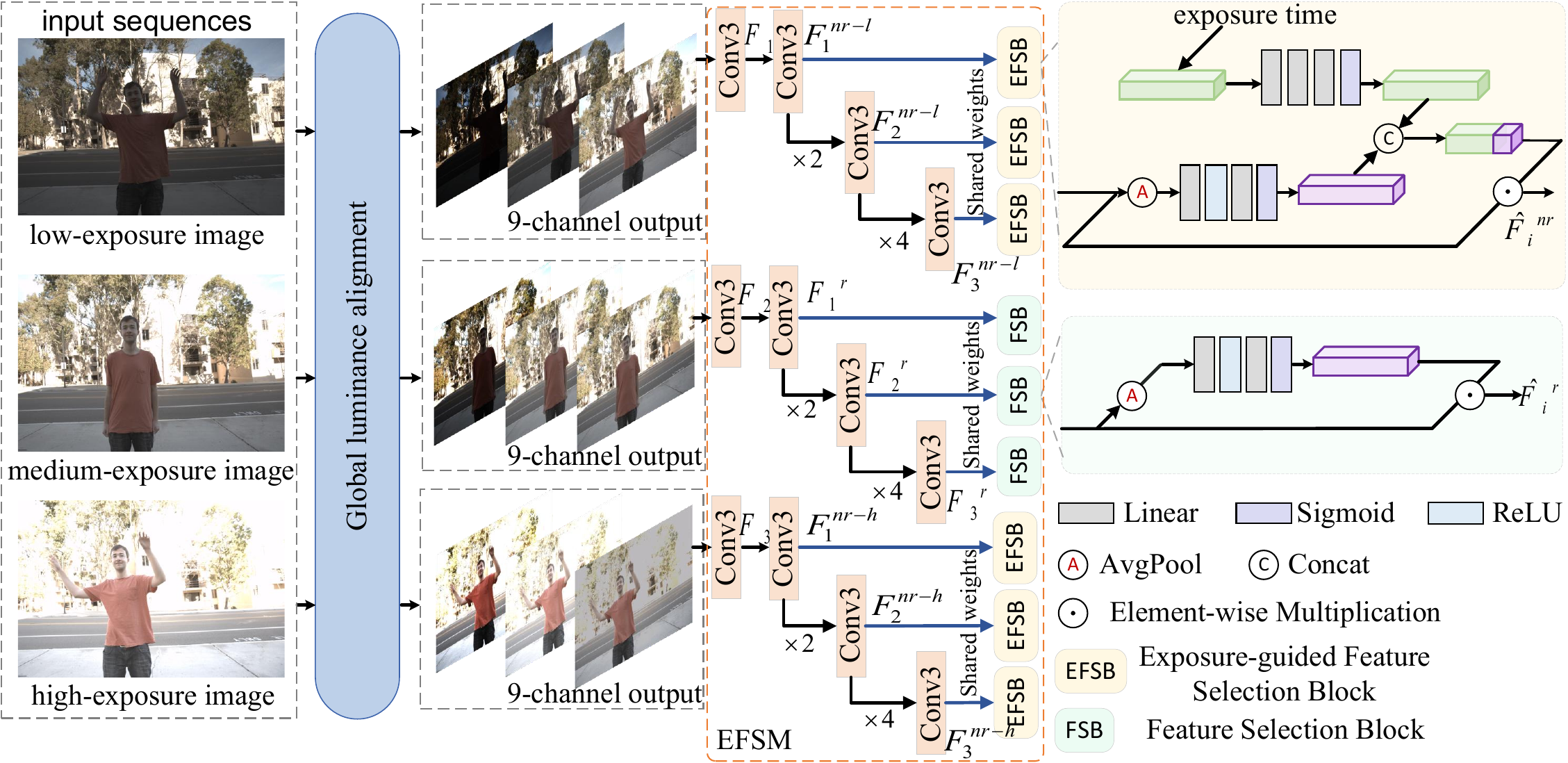} 
\caption{The pre-alignment subnetwork is composed of two parts: global luminance alignment (GLA) and exposure-guided feature selection module (EFSM). The input is divided into multiple scales using a $3\times3$ convolution, with a shared-weight feature selection block applied at each scale.} 
\label{prealign}
\end{figure}

As shown in Fig.~\ref{Fig2}, given a set of LDR images with varying exposure times, we construct the input tensor 
\(\mathbf{L}_{\text{in}} = \big( \{ L_l, G_l, A_l \}, \{ L_m, G_m, A_m \}, \{ L_h, G_h, A_h \} \big)\), 
where each triplet \((L_i, G_i, A_i)\) represents the original, gamma-corrected, and GLA-processed versions of the image under exposure setting \(i \in \{l, m, h\}\). This tensor is then passed to the EFSM for feature selection and alignment refinement.

\noindent \textbf{Exposure-guided Feature Selection Module (EFSM).} 
Exposure information is crucial for distinguishing the reliability and detail across different exposures. For the obtained multi-channel features, we propose the EFSM, which consists of the Exposure-guided Feature Selection Block (EFSB) and the Feature Selection Block (FSB). EFSM introduces exposure-aware modulation based on relative exposure priors, selectively emphasizing well-exposed regions and suppressing unreliable features, thereby ensuring more reliable feature representation for fusion.

Specifically, EFSM operates in a multi-scale feature refinement architecture, as depicted in Fig. \ref{prealign}. Given a 9-channel composite input tensor $\mathbf{L}_{\text{in}}$, we first extract three initial feature maps $\{\mathbf{F}_l,\mathbf{F}_m,\mathbf{F}_h\}$ via parallel $3 \times 3$ convolutional layers. For subsequent processing clarity, we formally define $\mathbf{F}^{nr} \triangleq \{\mathbf{F}_l,\mathbf{F}_h\}$ as non-reference features and $\mathbf{F}^{r} \triangleq \mathbf{F}_m$ as reference features. These are fed into two distinct but weight-shared selection blocks: EFSB for non-reference features and FSB for reference features.

To encode exposure priors, we normalize the exposure times \(\{ t_l, t_m, t_h \}\) relative to the middle exposure as:
\begin{equation}
e_l = \log_2\!\left( \frac{t_l}{t_m} \right) \cdot c, \quad
e_h = \log_2\!\left( \frac{t_h}{t_m} \right) \cdot c,
\end{equation}
where \( e_l \) and \( e_h \) represent the relative exposure embeddings for the low- and high-exposure branches, respectively, and \( c \) is a hyper-parameter empirically set to \(0.1\).

Both $\mathbf{F}^{r}$ and $\mathbf{F}^{nr}$ are processed through global average pooling followed by two fully-connected layers and a sigmoid activation, producing feature-based modulation vectors:$\mathbf{v}^{nr} \in \mathbb{R}^{1 \times 1 \times 64}$ and $\mathbf{v}^{r} \in \mathbb{R}^{1 \times 1 \times 64}$. In parallel, the exposure inputs ${e_l, e_h}$ are passed through a shared fully-connected layer to obtain exposure-based modulation vectors $\mathbf{v}^t\in \mathbb{R}^{1 \times 1 \times 128}$.
To combine feature reliability and exposure priors, we compute the exposure-guided modulation weights as:
\begin{equation}
\mathbf{u}^{\text{nr}} = \sigma\!\big( \operatorname{FC}_3( [\mathbf{v}^{\text{nr}}; \mathbf{v}^{\text{t}}] ) \big),
\end{equation}
where \([\cdot; \cdot]\) denotes channel-wise concatenation, 
\(\sigma(\cdot)\) is sigmoid activation, and \(\operatorname{FC}_3(\cdot)\) represents three fully-connected layers. Finally, the refined features are obtained as:
\begin{align}
\hat{\mathbf{F}}_i^{\text{r}} = \mathbf{F}_i^{\text{r}} \odot \mathbf{v}^{\text{r}}, \quad
\hat{\mathbf{F}}_i^{\text{nr}} = \mathbf{F}_i^{\text{nr}} \odot \mathbf{u}^{\text{nr}},
\end{align}
where \(\odot\) denotes element-wise multiplication.

\subsection{asymmetric cross-feature fusion subnetwork}
\label{Asymmetric Cross-feature Fusion subnetwork}


Leveraging non-reference details without ghosting is a critical challenge in dynamic scenes \cite{yan2024dynamic, liu2022ghost}. We propose an asymmetric cross-feature fusion subnetwork centered on the Asymmetric Cross-Attention (ACA) module. Instead of relying on rigid local alignment, ACA reformulates the problem as global feature retrieval to select and aggregate reliable features. By leveraging multi-scale window-based attention, it captures long-range dependencies, effectively handling significant spatial offsets. Furthermore, a reference-dominant mechanism strictly anchors the fusion to the reference stream, preventing artifact propagation.

 \begin{figure}[!t] 
\centering 
\includegraphics[width=0.49\textwidth]{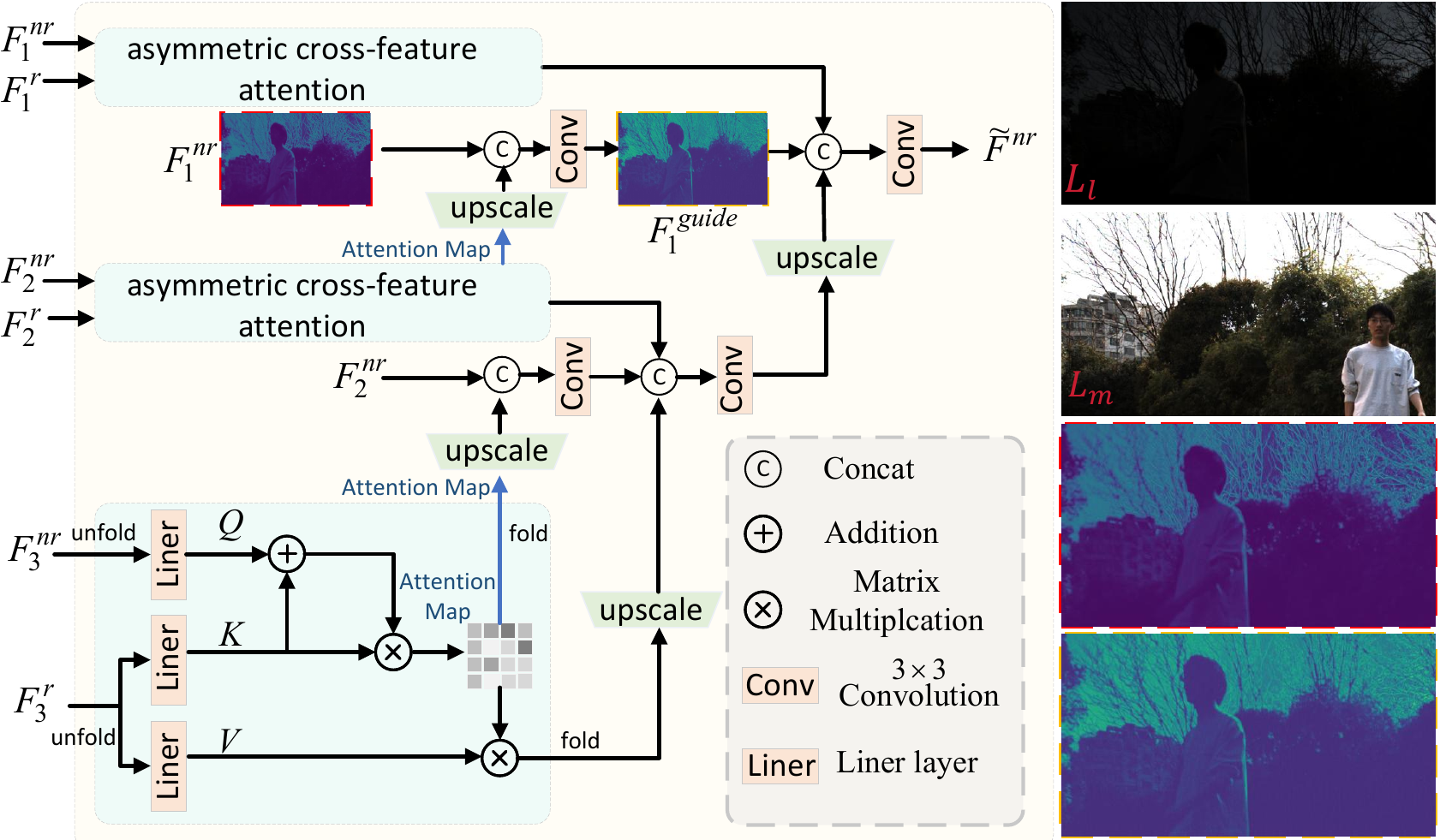} 
\caption{The structure of the asymmetric cross-feature fusion block. It employs asymmetric cross-feature attention to align reference and non-reference features, integrating coarse-to-fine guidance for enhanced reference-dominated feature fusion and improved HDR reconstruction.} 
\label{acfn} 
\end{figure}

As illustrated in Fig.~\ref{acfn}, the proposed module adopts a multi-scale architecture, where local attention at each level is computed via an ACA mechanism.
To mitigate feature misalignment caused by large motion, we propose an asymmetric attention mechanism that incorporates a structural prior into the matching process.
Let $\mathbf{F}^{\text{nr}}$ and $\mathbf{F}^{\text{r}}$ denote the non-reference and reference features, respectively. We formulate the query $\mathbf{Q}$ from the non-reference stream and the key $\mathbf{K}$ from the reference stream. Unlike standard cross-attention, we explicitly inject reference features into the query projection to regularize the attention distribution:
\begin{equation}
     \mathbf{M}_i = \operatorname{softmax}\!\left( (\mathbf{Q}_i+ \mathbf{K}_i) \mathbf{K}_i^\top/{\sqrt{C}} \right),
\end{equation}
where $\mathbf{Q}_i = \phi(\hat{\mathbf{F}}_i^{\text{nr}})\,\mathbf{W}_i^{q}$ and $\mathbf{K}_i,\mathbf{V}_i =\phi(\hat{\mathbf{F}}_i^{\text{r}})\,\mathbf{W}_i^{k,v}$, with $\mathbf{W}_i^{q, k,v}$ being learnable projection matrices. This formulation 
reveals an implicit competition mechanism between cross-correlation and self-correlation. In aligned regions, the cross-term dominates, capturing correspondence. However, in occluded or misaligned regions where $\mathbf{Q}$ and $\mathbf{K}$ diverge, the cross-term vanishes. Consequently, the self-correlation term dominates the attention landscape. This acts as a fallback regularization, anchoring the attention weights to the reliable reference structure and effectively suppressing ghosting artifacts arising from the non-reference stream.

Then, the aligned non-reference features $\mathbf{F}^{anr}_{i}$ are obtained by applying a folding operation to the result of the multiplication of $\mathbf{M}_i$ and $\mathbf{V}_i$.
To enhance fine-scale alignment with global context, we incorporate a cross-scale guidance mechanism. Specifically, attention information from the coarser scale $i+1$ is projected and injected into the finer scale $i$ as:
\begin{equation}
\mathbf{F}_i^{\text{guid}} = \operatorname{Conv}_{3\times 3}\big([\mathbf{F}_i^{\text{nr}}; U(\psi(\mathbf{M}_{i+1}))]\big), \quad 1 \leq i < N,
\end{equation}
where \(\operatorname{Conv}_{3\times 3}(\cdot)\) denotes a \(3\times3\) convolution, $N$ denotes the number of refinement scales, \(U(\cdot)\) is pixel shuffle upsampling, and \(\psi(\cdot)\) represents the projection of attention from the coarser scale. Then, \(\mathbf{F}_i^{\text{guid}}\) injects reference-dominated attention to facilitate multi-scale alignment.
The output of the ACA $\mathbf{F}^{\text{nr}}$ is computed by merging the aligned features from multiple scales, combining the finer-scale features with guidance from the coarser scales to produce a refined feature representation.

Finally, we generate a spatial attention (SA) map using \(\tilde{\mathbf{F}}^{\text{nr}}\) and \(\mathbf{F}^{\text{ref}}\) and get the fusion output.


\subsection{restoration subnetwork}
\label{restoration subnetwork}

As shown in Fig.~\ref{Fig2}, the restoration subnetwork is designed to suppress ghosting artifacts while preserving fine details across multiple scales.
Each scale begins with a Discrete Wavelet Transform (DWT) to decompose features into frequency subbands. At the lowest scale, where high-frequency details such as textures and residual motion artifacts are most prominent, we apply a Lightweight DomainPlus Block (LDPB) to perform frequency-specific corrections. The LDPB is adapted from~\cite{zheng2022domainplus} by removing dense connections to reduce complexity while retaining frequency modeling capacity.

In contrast, higher scales capture more global and low-frequency content. These layers are processed using a simple \(1\times1\) convolution after DWT, which is sufficient for coarse structure refinement without the need for full spectral filtering. All features are subsequently reconstructed using the Inverse Wavelet Transform (IWT).
The refined features from each scale are fused in a top-down manner using upsampling and residual connection to produce the final HDR output.

\subsection{Loss Function}
\label{lossfunction}
HDR images are typically displayed after tone mapping. To enhance training performance, we apply the $\mu$-law transformation, as proposed in \cite{kalantari1}, to map the HDR image from the linear domain to the tone-mapped domain, thereby improving model performance during training, the $\mu$-law transformation can be expressed as:
    \begin{equation}
    \tau(\mathcal{H})=\frac{\ln (1+\mu \mathcal{H})}{\ln (1+\mu)}
    \end{equation}
where $\mu$ is a hyperparameter set to 5000.

We adopt the L1 loss as the basic loss function for pixel-wise supervision. However, relying solely on L1 loss often treats pixels independently, leading to over-smoothed results where high-frequency details are lost. To address this issue and improve the visual appearance of the restored images, we adopt the dilated advanced Sobel loss (D-ASL) \cite{zheng2021learning} as an additional objective. The D-ASL utilizes dilated convolutions to expand the receptive field, allowing the network to capture multi-scale gradient information. This effectively suppresses blur artifacts and produces sharper edges and richer textures compared to the basic L1 loss. The function is defined as:
\begin{equation}
    \mathcal{A} \mathcal{S} \mathcal{L}(\widehat{Z}, Z)=\frac{1}{N} \sum \mid \operatorname{Sobel}^{\star}(\widehat{Z})-\text {Sobel}^{\star}(Z) \mid\\
\end{equation}
\begin{equation}
    {\cal D} - {\cal A}{\cal S}{{\cal L}^{\left\{ {{d_1},{d_2}, \cdots ,{d_n}} \right\}}} = \sum\limits_{i = 1}^n {{\cal A}{\cal S}{\cal L}\left| {_{dilation\_rate = {d_i}}} \right.} 
\end{equation}
where $N$ is the batch size of training, $Z$ and $\widehat{Z}$ represent the groundtruth and the output image predicted by the network after the $\mu$-law function, $\operatorname{Sobel}^{\star}$ denotes the advanced Sobel filtering, and ${{\cal A}{\cal S}{\cal L}\left| {_{dilation\_rate = {d_i}}} \right.}$ denotes the ASL with a dilation rate of $d_i$. Following \cite{zheng2021learning}, we apply the reference settings where $D = \left\{ {1,2,3} \right\}$ to formulate the D-ASL. Then the final loss function can be expressed as:
\begin{equation}
    \mathcal{L}_{total}=\mathcal{L}_1(\widehat{Z}, Z)+\lambda \cdot \mathcal{D}-\mathcal{A S} \mathcal{L}^D(\widehat{Z}, Z)
\end{equation}
where $\lambda$ is set to 0.25 to balance the L1 loss and D-ASL.

\begin{table*}[!t]
\centering
  \caption{Comparison results on Kalantari’s dataset and Prabhakar’s dataset. Throughout this paper, the best and second-best results of each case are highlighted in \best{bold red} and \secondbest{underlined blue}, respectively.}
\resizebox{1.0\textwidth}{!}{
\begin{tabular}{cc|ccccc|ccccc}
\shline
\rowcolor[HTML]{EFEFEF} 
\multicolumn{1}{c|}{\cellcolor[HTML]{EFEFEF}} & dataset setting & \multicolumn{5}{c|}{\cellcolor[HTML]{EFEFEF}train and test on Kalantari’s dataset} & \multicolumn{5}{c}{\cellcolor[HTML]{EFEFEF}train and test on Prabhakar’s dataset} \\ \hhline{>{\arrayrulecolor[HTML]{EFEFEF}}->{\arrayrulecolor{black}}|-----------} 
\rowcolor[HTML]{EFEFEF} 
\multicolumn{1}{c|}{\multirow{-2}{*}{\cellcolor[HTML]{EFEFEF}Method}} & metrics & PSNR-$\mu$ ($\uparrow$) & PSNR-PU ($\uparrow$) & SSIM-$\mu$ ($\uparrow$) & SSIM-PU ($\uparrow$) & HDR-VDP-2($\uparrow$) & PSNR-$\mu$ ($\uparrow$) & PSNR-PU ($\uparrow$) & SSIM-$\mu$ ($\uparrow$) & SSIM-PU ($\uparrow$) & HDR-VDP-2 ($\uparrow$) \\ \hline \hline
\multicolumn{2}{c|}{Kalantari(17' CGF) \cite{kalantari2017deep}} & 42.74 & - & 0.9877 & - & 60.51 & 35.63 & - & 0.9613 & - & 59.42 \\
\multicolumn{2}{c|}{AHDRNet(19' CVPR) \cite{yan2019attention}} & 43.77 & 36.97 & 0.9907 & 0.9833 & 62.30 & 38.61 & 30.88 & 0.9663 & 0.9319 & 61.14 \\
\multicolumn{2}{c|}{Prabhakar(20' ECCV ) \cite{prabhakar2020}} & 43.08 & - & - & - & 62.21 & 38.30 & - & 0.9702 & - & - \\
\multicolumn{2}{c|}{HDR-Trans(22' ECCV) \cite{liu2022ghost}} & 44.28 & \secondbest{37.50} & 0.9916 & \secondbest{0.9853} & 66.03 & \secondbest{41.31} & 34.60 & \secondbest{0.9726} & 0.9622 & \secondbest{63.01} \\
\multicolumn{2}{c|}{DomainPlus(22' MM) \cite{zheng2022domainplus}} & 44.02 & 37.15 & 0.9910 & 0.9841 & 62.91 & 40.38 & 33.86 & 0.9698 & 0.9608 & 62.12 \\
\multicolumn{2}{c|}{SCTNet(23' ICCV) \cite{tel2023alignment}} & 44.13 & 37.54 & 0.9916 & 0.9847 & 66.65 & 41.23 & \secondbest{34.63} & 0.9724 & \secondbest{0.9647} & 62.29 \\
\multicolumn{2}{c|}{SAFNet(24' ECCV) \cite{kong2024safnet}} & \secondbest{44.61} & 34.10 & \secondbest{0.9918} & 0.9796 & \secondbest{66.93} & 40.18 & 33.62 & 0.9705 & 0.9619 & 62.04 \\
\rowcolor[HTML]{FFEEED} 
\multicolumn{2}{c|}{\cellcolor[HTML]{FFEEED}EAFNet(Ours)} & \best{44.69} & \best{37.79} & \best{0.9920} & \best{0.9857} & \best{68.35} & \best{41.80} & \best{35.29} & \best{0.9731} & \best{0.9660} & \best{63.53} \\ \shline
\end{tabular}}
\label{tab5}
\end{table*}

\begin{table*}[]
\centering
  \caption{Cross-dataset evaluation on Kalantari’s dataset and Prabhakar’s dataset.}
  \resizebox{\linewidth}{!}{
\begin{tabular}{cc|cccc|cccc}
\hline
\rowcolor[HTML]{EFEFEF} 
\multicolumn{1}{c|}{\cellcolor[HTML]{EFEFEF}} & dataset setting & \multicolumn{4}{c|}{\cellcolor[HTML]{EFEFEF}train on Kalantari’s dataset, test on Prabhakar’s dataset} & \multicolumn{4}{c}{\cellcolor[HTML]{EFEFEF}train on Prabhakar’s dataset, test on Kalantari’s dataset} \\ \hhline{>{\arrayrulecolor[HTML]{EFEFEF}}->{\arrayrulecolor{black}}|---------}  
\rowcolor[HTML]{EFEFEF} 
\multicolumn{1}{c|}{\multirow{-2}{*}{\cellcolor[HTML]{EFEFEF}Method}} & metrics & PSNR-$\mu$ ($\uparrow$) & PSNR-PU ($\uparrow$)  & SSIM-$\mu$ ($\uparrow$) & SSIM-PU ($\uparrow$) & PSNR-$\mu$ ($\uparrow$) & PSNR-PU ($\uparrow$) & SSIM-$\mu$ ($\uparrow$) & SSIM-PU ($\uparrow$) \\ \hline \hline
\multicolumn{2}{c|}{AHDRNet(19' CVPR) \cite{yan2019attention}} & 33.96 & 26.89 & 0.9601 & 0.9273 & 40.03 & 32.25 & 0.9855 & 0.9453 \\
\multicolumn{2}{c|}{HDR-Trans(22' ECCV) \cite{liu2022ghost}} & 34.07 & 30.60 & \secondbest{0.9675} & \secondbest{0.9429} & \secondbest{41.38} & 32.70 & 0.9890 & 0.9623 \\
\multicolumn{2}{c|}{DomainPlus(22' MM) \cite{zheng2022domainplus}} & 32.64 & 27.74 & 0.9026 & 0.9417 & 41.15 & 32.82 & 0.9873 & \secondbest{0.9679} \\
\multicolumn{2}{c|}{SCTNet(23' ICCV) \cite{tel2023alignment}} & 33.83 & 27.35 & 0.9584 & 0.9427 & 40.88 & \secondbest{33.05} & \secondbest{0.9892} & 0.9611 \\
\multicolumn{2}{c|}{SAFNet(24' ECCV) \cite{kong2024safnet}} & \secondbest{38.00} & \secondbest{30.61} & 0.9597 & 0.9264 & 40.86 & 28.93 & 0.9882 & 0.7741 \\
\rowcolor[HTML]{FFEEED} 
\multicolumn{2}{c|}{\cellcolor[HTML]{FFEEED}EAFNet(Ours)} & \best{39.26} & \best{32.62} & \best{0.9707} & \best{0.9625} & \best{42.02} & \best{33.08} & \best{0.9903} & \best{0.9692} \\ \hline
\end{tabular}}
\label{tabcross}
\end{table*}

\section{EXPERIMENTS}
\label{experiment}


We conduct extensive experiments to evaluate our dual-stream paradigm. Section~\ref{Experiment Settings} outlines the experimental setup. Section~\ref{HDR Image Experimental Results} benchmarks our method against state-of-the-art deghosting approaches, validating its single-frame fusion capability. Section~\ref{expvideo} focuses on video reconstruction, demonstrating superior temporal consistency and flicker suppression over AE paradigms. We then analyze component contributions in Section~\ref{Ablation study} and investigate parallax robustness in Section~\ref{Parallax Analysis}. Finally, Section~\ref{limitation} discusses limitations and future directions.

\subsection{Experiment Settings}
\label{Experiment Settings}

\noindent \textbf{Training Details.} The proposed method is implemented in PyTorch on an NVIDIA RTX4090 GPU. For the convolutional layers, we employ 64 kernels of size 3$\times$3 with a stride of one, applying zero padding to preserve the dimensions of the resulting feature maps. The optimization is performed using the Adam optimizer \cite{kingma2014adam} with an initial learning rate of $10^{-4}$, and the training concludes when the learning rate reaches $10^{-6}$, which typically requires approximately 72 hours.
Both the LDR and corresponding HDR images are cropped into 256 $\times$ 256 patches. Patches are randomly rotated to avoid overfitting. The batch size is set to 16.

\begin{figure*}[t] 
\centering 
\includegraphics[width=0.95\textwidth]{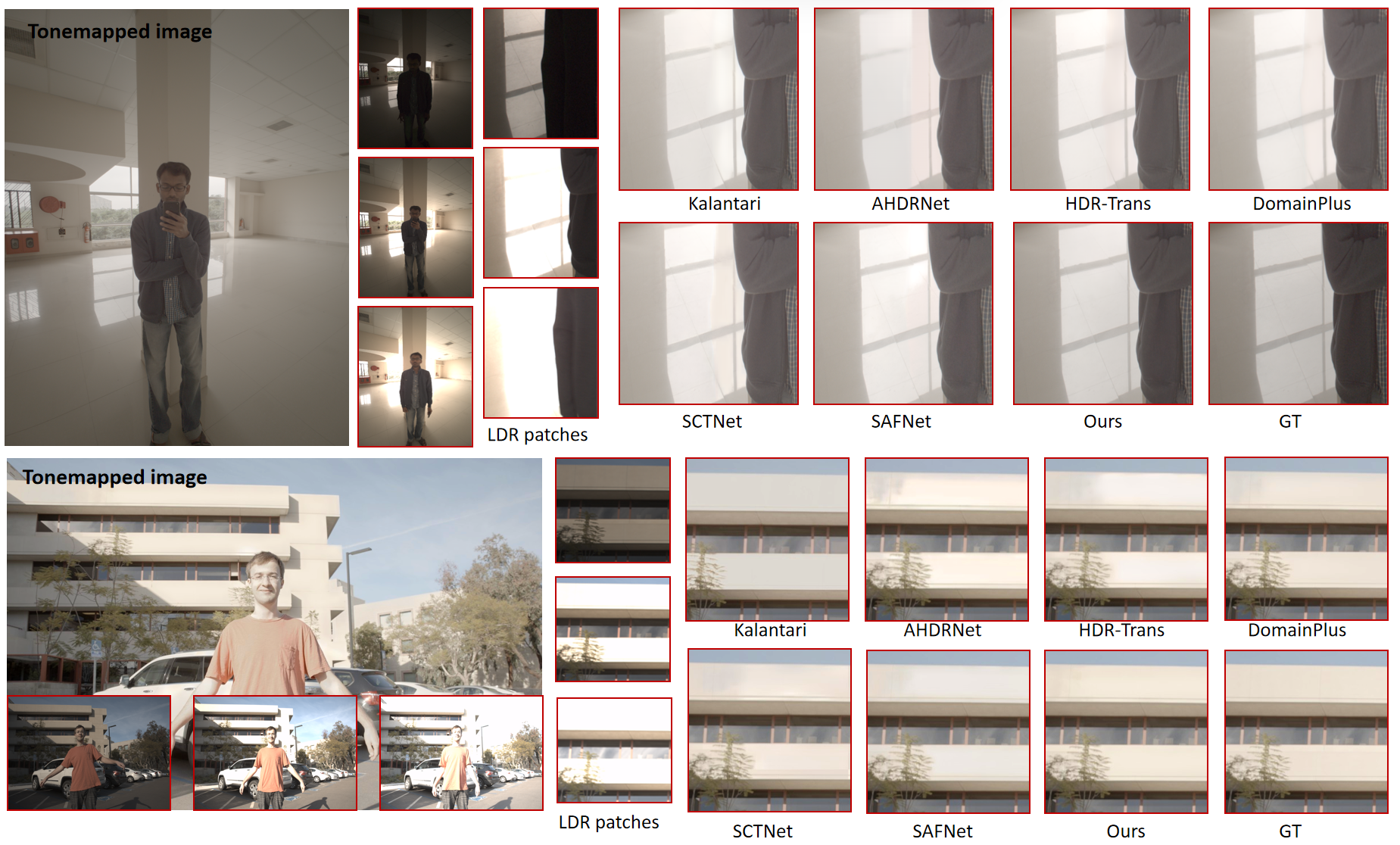} 
\caption{Visual comparison on Kalantari's and Prabhakar's dataset. While other methods suffer from detail loss, overexposed patches, and color inconsistencies, our approach successfully reconstructs overexposed details and preserves natural, consistent color.} 
\label{kalashow} 
\end{figure*}

\noindent \textbf{Datasets.}
High-quality HDR video datasets suitable for supervised training or frame-wise quantitative evaluation are scarce. Although a recent large-scale benchmark \cite{shu2024towards} has been introduced to facilitate real-world HDR video evaluation, public resources with dense per-frame HDR ground truth and diverse exposure conditions remain limited. 
To fairly and comprehensively evaluate our method, we separate the evaluation into two complementary parts:
\begin{itemize}
\item \textbf{Spatial evaluation} focuses on single-frame fusion accuracy using HDR image datasets with reliable ground truth, enabling objective comparison in reconstruction quality. We assess single-frame fusion quality using two widely used HDR image datasets: Kalantari’s dataset~\cite{kalantari1} and Prabhakar’s dataset~\cite{prabhakar2019fast}. 
\textit{Kalantari's dataset} includes 74 training samples and 15 testing samples, with exposure values of \{-2 EV, 0 EV, +2 EV\} or \{-3 EV, 0 EV, +3 EV\}.
\textit{Prabhakar's Dataset} includes 466 training samples and 116 testing samples, with exposure values between -3EV and +3EV. 

\item \textbf{Temporal evaluation} focuses on luminance consistency and flicker suppression in dynamic scenes, which cannot be assessed using static image datasets.
To ensure a rigorous and comprehensive evaluation, we conduct experiments using two distinct categories of data: 

\textit{Public Video Benchmarks (Synthetic Data).} Adopting the synthesis protocols from~\cite{xu2024hdrflow, chung2023lan, chen2021hdr}, we generated a large-scale HDR video training set based on the Vimeo-90K dataset~\cite{xue2019video}. This facilitates the retraining of image-based baselines, ensuring a fair comparison with video-based methods. All evaluations are conducted on the publicly available Cinematic Video Dataset~\cite{chen2021hdr}.

\textit{Self-captured Dual-Camera Test Dataset (Real-world Data).} To validate our approach in practical scenarios, we introduced a dataset of 26 sequences captured under diverse real-world conditions. The data includes indoor/outdoor environments and day/night transitions, specifically targeting challenging cases like high-contrast motion and occlusions. Exposure settings are fixed to $(-2, 0, +2)$ or $(-1, 0, +1)$ EV to ensure sufficient exposure diversity across varying lighting conditions. 
By utilizing a dual-camera setup, we provide standard alternating exposure inputs for baselines while supplying the additional medium-exposure stream for our paradigm, ensuring a fair evaluation on identical scene content.

\end{itemize}   

\noindent \textbf{Baselines.}
We compare our EAFNet to representative state-of-the-art approaches, categorized into two groups:
\textit{Multi-Exposure HDR Deghosting Methods:} We select Kalantari~\cite{kalantari2017deep}, AHDR~\cite{yan2019attention}, Prabhakar~\cite{prabhakar2020}, HDR-Transformer~\cite{liu2022ghost}, SCTNet~\cite{tel2023alignment}, DomainPlus~\cite{zheng2022domainplus}, AFUNet~\cite{li2025afunet} and SAFNet~\cite{kong2024safnet}. 
\textit{AE-based HDR video Methods:} We compare with HDRFlow~\cite{xu2024hdrflow}, LAN-HDR~\cite{chung2023lan}, and DeepHDRVideo~\cite{chen2021hdr}.
For fairness, we exclude few-shot and unsupervised methods that are not designed for supervised multi-exposure HDR reconstruction.





\noindent \textbf{Metrics.}
We quantitatively evaluate the predicted HDR images by measuring PSNR and SSIM~\cite{wang2004image}. 
Following established protocols, we report these metrics in two domains: the tone-mapped domain (*-$\mu$)~\cite{kalantari1} and the PU-encoded domain (*-PU) ~\cite{azimi2021pu21}, which is perceptually uniform and better reflects human visual response to high dynamic range content.
The tone mapping operation used for evaluation is described in Sec.~\ref {lossfunction}. 
Additionally, we compare the results using HDR-VDP-2~\cite{mantiuk2011hdr} for conforming the human visual perception.

To comprehensively assess video quality and temporal stability, we further employ LPIPS and temporal metrics including t-PSNR, t-SSIM~\cite{wang2012spatio}, CVVDP~\cite{mantiuk2024colorvideovdp}, and FVVDP~\cite{mantiuk2021fovvideovdp}.
Specifically, to quantify luminance consistency and flicker, we report Luminance Standard Deviation (LSD),  standard deviation of frame-wise quality metrics (STD) and Mean Absolute Difference of Brightness (MADB).

\begin{table*}[]
\caption{Quantitative comparisons of our method with other state-of-the-art methods on the Cinematic Video dataset. The values are reported in the format \textbf{Slow / Fast}, representing performance on slow-motion and fast-motion scenes, respectively.
%
%
}
\resizebox{1.0\textwidth}{!}{
\begin{tabular}{c|cccccccccc}
\hline
\rowcolor[HTML]{EFEFEF} 
Methods             & \textbf{STD ($\downarrow$)} & \textbf{PSNR\_T ($\uparrow$)} & \textbf{SSIM\_T ($\uparrow$)} & \textbf{HDR-VDP-2 ($\uparrow$)} & \textbf{t-SSIM ($\uparrow$)} & \textbf{t-PSNR ($\uparrow$)}  & \textbf{MADB ($\downarrow$)} & \textbf{CVVDP ($\uparrow$)} & \textbf{FVVDP ($\uparrow$)} & \textbf{LPIPS ($\downarrow$)}\\ \hline
DeepHDRVideo (21' ICCV)     & 2.70/1.12                   & 35.60/29.60                   & 0.8802/0.8603                 & \best{66.76}/59.44                     & 0.6164/0.8092                & 38.79/34.08                                  & 1.22/4.30                    & 7.22/8.02                   & 6.67/7.89                   &                             0.3016/0.1233\\
LAN-HDR (23' ICCV)  & 2.80/\secondbest{1.07}                   & 35.75/\best{31.18}                   & 0.8505/0.8536                 & \secondbest{66.53}/57.85                     & 0.6300/0.8196                & 39.12/\best{34.59}                                  & 0.90/3.07                    & 7.71/8.16                   & 6.95/8.00                   &                             0.2955/0.0938\\
HDRFlow (24' CVPR)  & 0.90/\best{1.01}                   & 39.39/29.96                   & 0.9000/\best{0.8837}                 & 66.37/59.37                     & 0.6813/\best{0.8676}                & 42.77/33.16                        & 0.45/3.83                    & 9.11/8.31                   & 8.21/\best{8.08}                   &                            0.0514/\best{0.0478}\\  \hline
AHDNet (19’ CVPR)   & 0.49/1.14                   & 39.68/30.27                   & 0.9373/0.8533                 & 63.56/61.58                     & 0.7916/0.8358                & 43.86/33.57                                 & 0.12/\secondbest{1.80}                    & 8.99/8.34                   & 7.64/7.97                   &                             0.2028/0.1242\\
HDRTrans (22’ ECCV) & \secondbest{0.42}/1.25                   & \secondbest{40.32}/30.55                   & 0.9412/0.8676                 & 62.95/62.05                     & 0.8086/0.8464                & 44.19/33.80                                  & 0.17/1.81                    & 9.13/8.34                   & 7.92/8.00                   &                             0.1036/0.1004\\
DomainPlus (22’ MM) & 0.47/1.12                   & 40.14/30.51                   & 0.9406/0.8665                 & 63.39/63.06                     & 0.8379/0.8458                & \secondbest{44.42}/33.79                                 & 0.19/1.81                    & \secondbest{9.19}/8.34                   & \secondbest{8.32}/8.00                   &                             0.0959/0.1005\\
SCTNet (23’ ICCV)   & 0.45/1.17                   & 39.78/30.56                   & 0.9317/0.8752                 & 62.61/61.93                     & \secondbest{0.8416}/0.8498                & 44.29/33.83                                & 0.12/1.79                    & \secondbest{9.19}/\secondbest{8.35}                   & 8.56/7.95                   &                             \secondbest{0.0330}/\secondbest{0.0554}\\
AFUNet (25' ICCV)   & 0.44/1.20                   & 40.17/30.21                   & 0.9391/0.8663                 & 62.89/62.69                     & 0.8353/0.8483                & 44.31/33.83                              & \best{0.10}/1.92                    & 9.14/8.33                   & 7.96/7.98                   &                             0.0392/0.0768\\ \hline
\rowcolor[HTML]{FFEEED} 
Ours                & \best{0.39}/1.23                   & \best{40.87}/\secondbest{31.12}                   & \best{0.9456}/\secondbest{0.8826}                 & 63.76/\best{63.14}                     & \best{0.8650}/\secondbest{0.8604}                & \best{44.95}/\secondbest{34.30}                                  & \secondbest{0.11}/\best{1.79}                    & \best{9.32/8.44}                   & \best{8.56}/\secondbest{8.03}                   &                             \best{0.0179}/0.0624\\ \hline
\end{tabular}
}
\label{table:cin}
\end{table*}

\begin{table}[]
   \centering
  \caption{Comparison on self-captured videos. $*$ indicates HDR image deghosting methods under the alternating exposure solution. The reported time refers to the inference time for a 128×128 patch on a single RTX 4090 GPU.}
  \resizebox{0.5\textwidth}{!}{
\begin{tabular}{c|c|cccc}
\hline
\rowcolor[HTML]{EFEFEF} 
Method                                                          & Paradigm                                                                                       & LSD ($\downarrow$)                   & t-SSIM ($\uparrow$)                  & $L_{avg}$ & Inference time (ms) \\ \hline
DeepHDRVideo \cite{chen2021hdr}     &                                                                                               & 0.1031                               & 0.5913                               & 105.02    & 22.23                              \\
LAN-HDR \cite{chung2023lan}         &                                                                                               & 0.1003                               & 0.6879                               & 108.75    & 21.69                              \\
HDRFlow \cite{xu2024hdrflow}        &                                                                                               & 0.0914                               & 0.7352                               & 113.49    & \best{2.62}       \\
SCTNet* \cite{tel2023alignment}     &                                                                                               & 0.1027                               & 0.6947                               & 109.23    & 11.17                              \\
SAFNet* \cite{kong2024safnet}       & \multirow{-5}{*}{\begin{tabular}[c]{@{}c@{}}alternating-\\ exposure\\  paradigm\end{tabular}} & 0.0976                               & 0.6862                               & 111.70    & 3.27                               \\ \hline
HDR-Trans \cite{liu2022ghost}       &                                                                                               & \best{0.0093}       & 0.8923                               & 111.24    & 21.60                              \\
DomainPlus \cite{zheng2022domainplus} &                                                                                               & \secondbest{0.0094} & \secondbest{0.8991} & 110.41    & 7.44                               \\
SCTNet \cite{tel2023alignment}      &                                                                                               & \best{0.0093}       & 0.8868                               & 106.84    & 11.17                              \\
SAFNet \cite{kong2024safnet}        & \multirow{-4}{*}{\begin{tabular}[c]{@{}c@{}} dual-stream \\ paradigm \\ (Ours)\end{tabular}}      & 0.0097                               & 0.8851                               & 111.32    & \secondbest{3.24} \\
\rowcolor[HTML]{FFEEED} 
EAFNet(Ours)                                                    & \multicolumn{1}{l|}{\cellcolor[HTML]{FFEEED}}                                                 & \best{0.0093}       & \best{0.9071}       & 133.16    & 4.32                               \\ \hline
\end{tabular}
}
 \label{videotable}
\end{table}

\subsection{HDR Image Experimental Results}
\label{HDR Image Experimental Results}

\noindent \textbf{Quantitative Comparisons.}
Table~\ref{tab5} reports the intra-dataset evaluation results on Prabhakar’s and Kalantari’s datasets. Our EAFNet consistently achieves the best performance across both datasets. On the Kalantari dataset, it surpasses the second-best method by $0.08$ dB in PSNR-$\mu$, whereas on the Prabhakar’s dataset, which offers a significantly larger and more diverse evaluation set, the margin increases to $0.49$ dB. 

We further conduct cross-dataset validation, which is critical for evaluating the robustness of HDR fusion models against variations in exposure patterns, motion distributions, and scene content. As shown in Table~\ref{tabcross}, our EAFNet maintains clear superiority in this challenging setting, with cross-domain gains even larger than in the intra-dataset case. For example, when trained on Kalantari and tested on Prabhakar, EAFNet achieves 39.26 dB PSNR-$\mu$ and 0.9707 SSIM-$\mu$, surpassing the next best approach by over 1.2 dB and 0.03, respectively. This demonstrates that our model does not overfit to dataset-specific statistics, but learns exposure-aware and motion-robust fusion representations that transfer effectively across domains. The strong bidirectional results confirm the generality and domain-agnostic nature of our fusion mechanism.

\noindent \textbf{Qualitative Comparisons.}
Fig.~\ref{kalashow} presents visual comparisons on the Prabhakar and Kalantari datasets. Most existing methods exhibit noticeable ghosting artifacts, particularly around moving objects and edges with exposure differences. In overexposed regions, the Kalantari method suffers from detail loss, while AHDRNet and HDR-Trans produce large saturated patches, and DomainPlus struggles with color inconsistencies. In contrast, EAFNet recovers fine structures in saturated areas while preserving natural and consistent color reproduction.


    \begin{figure*}[t] 
    \centering 
    \includegraphics[width=0.97\textwidth]{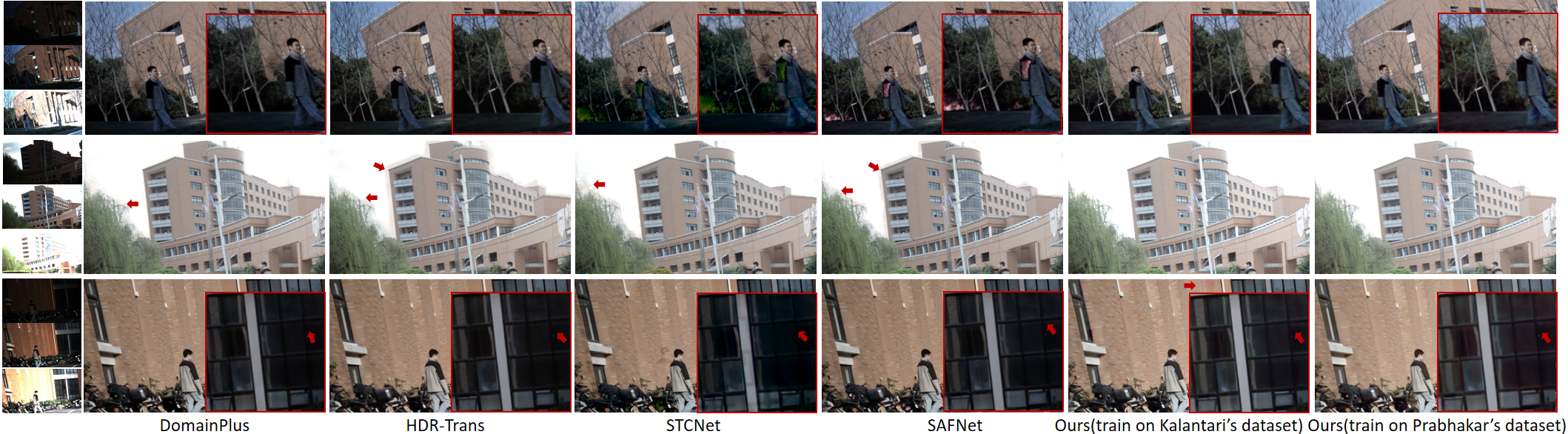} 
    \caption{Visual comparison with HDR image deghosting methods on self-captured videos. Our EAFNet reduces ghosting from motion (first row) and background misalignment (second row), recovering natural colors and fine details in dark regions (third row).} 
    \label{videomef} 
    \end{figure*}    

\begin{figure}[t] 
\centering 
\includegraphics[width=0.49\textwidth]{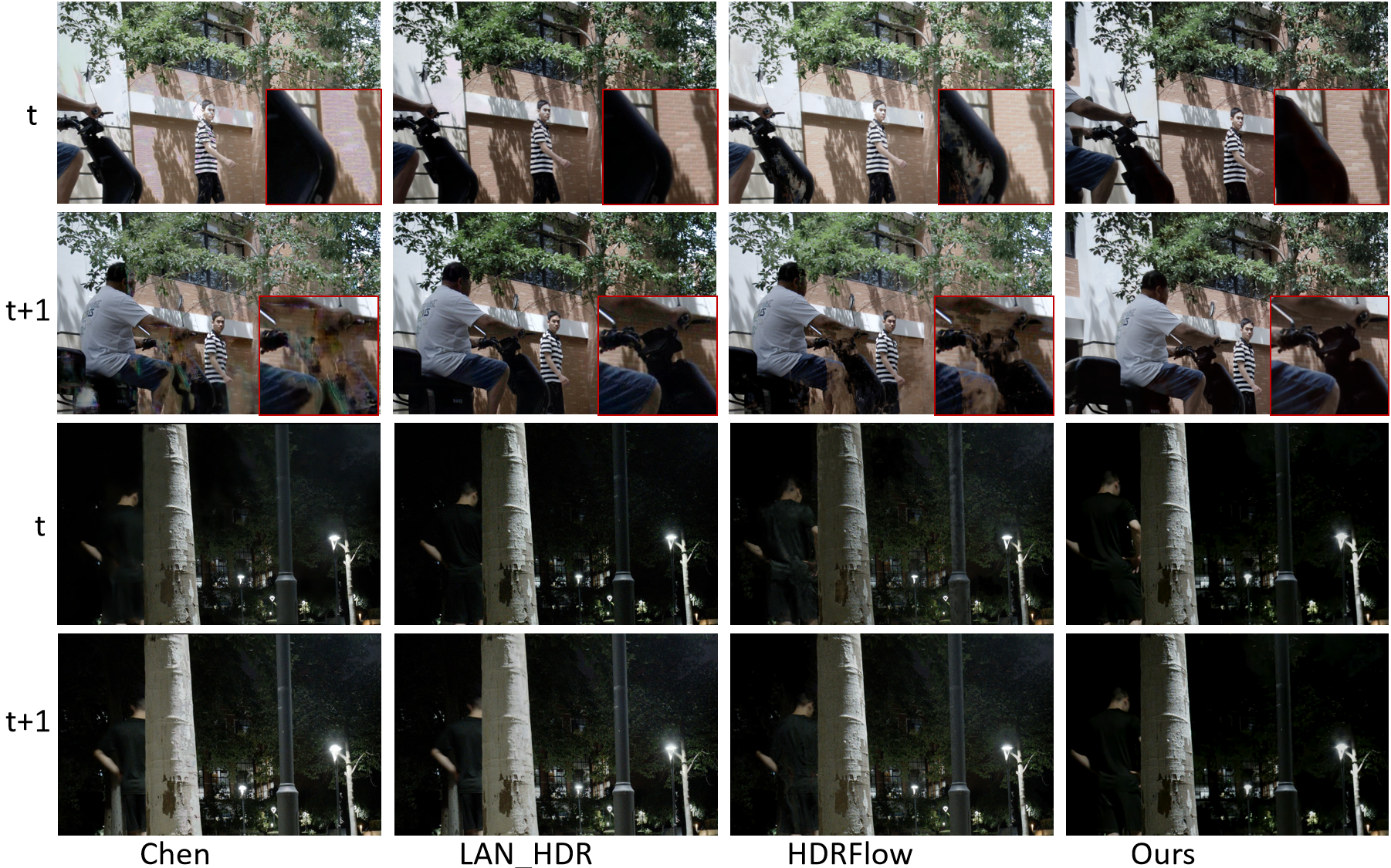} 
\caption{Visual comparison with AE methods on self-captured video dataset.
} 
\label{video_vis_ae} 
\end{figure}

\subsection{HDR Video Experimental Results}
\label{expvideo}

\noindent \textbf{Paradigm-level comparison.
} 
The quantitative superiority shown in Table~\ref{table:cin} and Table~\ref{videotable}  (e.g., lower MADB and higher t-PSNR) underscores a fundamental architectural advantage of our dual-stream paradigm over AE-based pipelines.
Furthermore, as shown in Fig.~\ref{video_vis_ae}, AE-based reconstructions suffer from severe artifacts in fast-motion scenarios: the substantial motion between alternating exposures leads to structural tearing, while the inconsistency in reference frames causes noticeable luminance fluctuations between consecutive frames.

The root cause of this performance gap lies in the stability of the reference anchor:
AE methods (e.g., HDRFlow, LAN-HDR) rely on an alternating sequence. Consequently, the reference frame itself suffers from constant photometric jumping. The network must struggle to distinguish whether a pixel's intensity change is due to scene motion or exposure switching, leading to inevitable flickering and high STD/MADB values.
In contrast, our paradigm decouples temporal stability from HDR fusion. By maintaining a continuous, fixed-exposure reference stream, we provide a luminance invariant backbone for the video. The reconstruction process is reduced to enhancing a stable video rather than stabilizing a jumping sequence. This structural guarantee allows our architecture to consistently minimize temporal fluctuations, outperforming AE baselines.

\noindent \textbf{ Within-dual-stream comparison.}
We further compare EAFNet against state-of-the-art image deghosting baselines operating under the same dual-stream framework. Note that for the Cinematic benchmark, these baselines were retrained on Vimeo-90K for fair comparison.

As shown in Fig.~\ref{videomef},
in high-motion scenarios, baselines (e.g., DomainPlus, SAFNet) often generate ghosting artifacts due to their inability to filter out misaligned features. In contrast, our ACA module effectively suppresses these parallax-induced outliers, preserving sharp spatial structures.
Furthermore, under extreme luminance contrast, baselines introduce color distortions or noise in dark regions. Here, our EFSM demonstrates its robustness to exposure variation, recovering fine structures in localized low-light regions and maintaining natural transitions in overexposed areas without introducing noise.
These visual improvements align with the quantitative margins in Table~\ref{videotable}, confirming that our specialized modules handle the dual challenges of exposure disparity and spatial misalignment more effectively than generic baselines.

\begin{figure*}[!t] 
\centering     \includegraphics[width=0.97\textwidth]{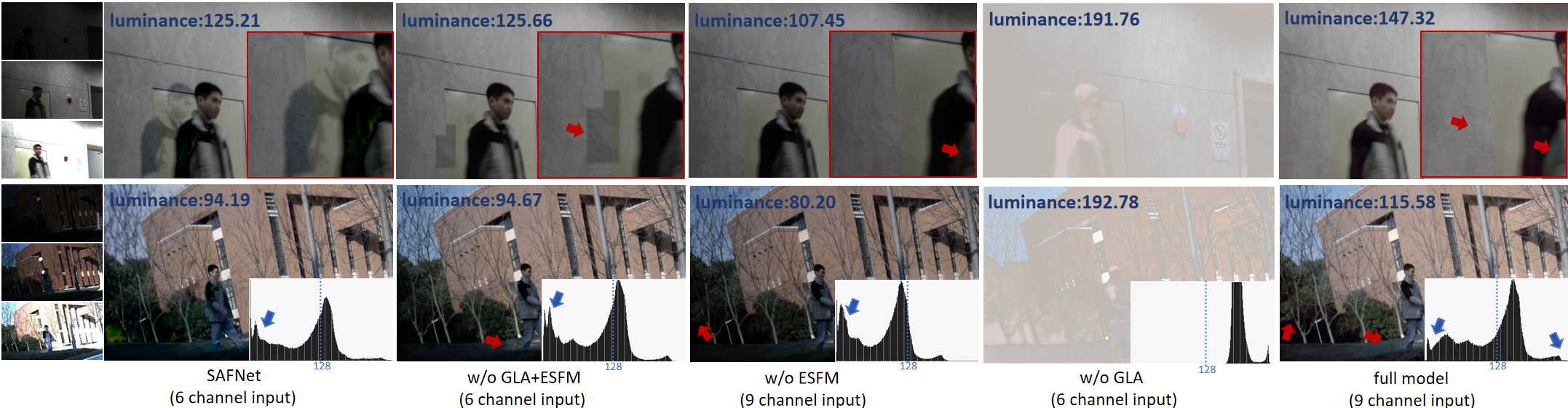} 
\caption{Visual comparison of ablation on self-captured videos. The integration of GLA and EFSM influences the overall luminance and jointly reduces ghosting while enhancing detail in dark regions. The luminance histograms show that our method increases the visibility of extremely dark areas without causing an increase in high-luminance pixels.} 
\label{abpre} 
 \end{figure*}   

\begin{figure}[t] 
\centering     \includegraphics[width=0.45\textwidth]{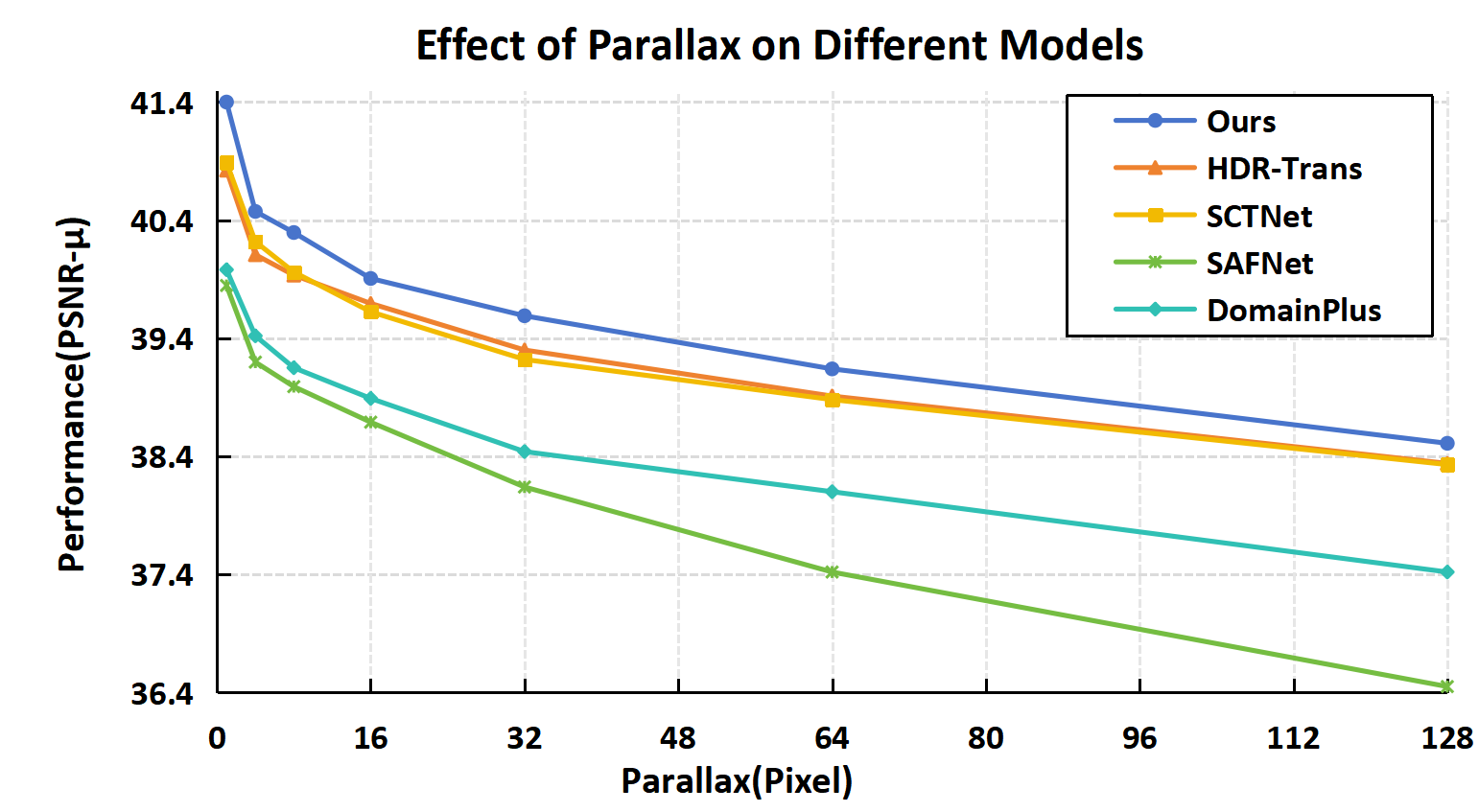} 
\caption{Performance variation of different models under simulated parallax.} 
\label{parallax} 
\end{figure}   

\subsection{Parallax Analysis}
\label{Parallax Analysis}

\noindent \textbf{Quantification of the Parallax Challenge.} The fixed-baseline configuration of our dual-camera system inherently introduces disparity between views. Considering our hardware setup ($B \approx 44\mathrm{mm}$ and effective focal length due to inference resizing), the disparity for a target at distance $Z$ follows the geometric model $\Delta x \approx (f_{\mathrm{px}}B)/{Z}$. Consequently, for close-range objects (e.g., $0.4\mathrm{m}$), the disparity can reach approximately $100$ pixels. 
This poses a significant challenge for dual-camera fusion algorithms.

\noindent \textbf{Comparative Stress Test.} To evaluate robustness against this theoretical maximum, we conducted a controlled stress test using the 116 samples from Prabhakar's test set. We synthetically introduced varying degrees of horizontal disparity (ranging from 0 to 100 pixels) to simulate the calculated parallax levels, comparing our method against adapted deep learning-based baselines (e.g., SAFNet, DomainPlus). As shown in Fig.~\ref{parallax}, the performance of all methods naturally degrades as parallax increases (0 to 100 pixels) due to occlusion and information loss. However, a distinct trend is observed: 
CNN-based Baselines (Green Lines): Methods like SAFNet and DomainPlus exhibit a sharp drop in PSNR. These approaches rely on local receptive fields, which struggle to capture such large displacements (100 px), leading to severe ghosting artifacts when alignment fails. 
Transformer-based baselines (Yellow Lines): Methods like HDR-Trans and SCTNet perform better than CNNs. Thanks to the global receptive field inherent to Transformers, they can theoretically capture long-range dependencies and retrieve features across large spatial distances.
Ours (Blue Line): Our method maintains the highest stability across the entire disparity range. This superior robustness stems from our unique Reference-Based Selective Fusion mechanism. Our Asymmetric Cross-Attention treats large parallax as a selection problem: When parallax is too large (e.g., occlusion), the attention module assigns low weights to the auxiliary view, effectively rejecting misaligned features.
Consequently, even under extreme parallax, our failure mode is merely a local loss of dynamic range enhancement rather than the catastrophic structural destruction seen in AE-based methods.

It is worth noting that AE methods are excluded from this specific parallax analysis. Since AE methods operate on a monocular sequence, they do not suffer from spatial disparity. However, as discussed in Sec.~\ref{expvideo}, their trade-off is extreme sensitivity to temporal motion and reference instability.
Our system accepts the challenge of spatial parallax, which we solve robustly as shown above, in exchange for solving the more critical issue of temporal flickering.


\subsection{Ablation study}
\label{Ablation study}

In this section, we first evaluate the robustness of our framework under different dual-camera deployment conditions using our self-captured real-world dataset, focusing on the impact of calibration accuracy, input ordering, and frame rate. 
We then perform an ablation study on EAFNet to analyze the contribution of each component and the source of efficiency, with all results reported on Prabhakar’s dataset~\cite{prabhakar2019fast}.

\noindent \textbf{Dual-Camera Analysis.}
In a dual-camera system, imperfect calibration and potential desynchronization can lead to geometric misalignment and motion-induced misalignment, respectively. In our prototype, no special effort is made to achieve sub-pixel calibration or perfect synchronization. Instead, the fusion framework is designed to tolerate moderate deviations in both geometry and timing. The training data include dynamic scenes with motion and occlusion, allowing the model to learn robustness to such spatial inconsistencies.

Table~\ref{dua-camera} reports t-SSIM under four deployment variants. Variations across deployment variants are small for all methods ($\leq$0.003), indicating that, within the tested range, moderate calibration errors, input order changes, and frame-rate reductions have limited impact on temporal flickering. Minor fluctuations are likely due in part to the cropping required for alignment rather than alignment failure itself. Overall, the results suggest that, within the tested range of deviations, temporal stability can be maintained without the need for precise calibration and strict synchronization, but a higher level of calibration and synchronization may still be beneficial for optimizing performance in more demanding scenarios.

\begin{table}[]
  \centering
  \caption{t-SSIM comparison under different dual-camera deployment settings. \textcolor{blue}{base:} Baseline dual-camera system settings. \textcolor{blue}{w/o Cal.:} Dual-camera system without calibration. \textcolor{blue}{Random nr.:} Randomized input order of non-reference frames.
\textcolor{blue}{Lower fr.:} Lower frame rate for non-reference frame input. }
\resizebox{0.5\textwidth}{!}{
\begin{tabular}{cc|cccc}
\hline
\rowcolor[HTML]{EFEFEF} 
\multicolumn{1}{c|}{\cellcolor[HTML]{EFEFEF}Method} & settings & Base. & w/o Cal. & Random nr. & Lower fr. \\ \hline \hline
\multicolumn{2}{c|}{HDR-Trans(ECCV 2022) \cite{liu2022ghost}} & 0.8923 & 0.8922 & 0.8926 & 0.8924 \\
\multicolumn{2}{c|}{DomainPlus(MM 2022) \cite{zheng2022domainplus}} & 0.8991 & 0.8994 & 0.8997 & 0.8985 \\
\multicolumn{2}{c|}{SCTNet(ICCV 2023) \cite{tel2023alignment}} & 0.8868 & 0.8872 & 0.8842 & 0.8871 \\
\multicolumn{2}{c|}{SAFNet(ECCV 2024) \cite{kong2024safnet}} & 0.8851 & 0.8861 & 0.8848 & 0.8836 \\
\rowcolor[HTML]{FFEEED} 
\multicolumn{2}{c|}{\cellcolor[HTML]{FFEEED}EAFNet(Ours)} & 0.9071 & 0.9079 & 0.9072 & 0.9066 \\ \hline
\end{tabular}}
\label{dua-camera}
\end{table}

\begin{table}[]
  \centering
  \caption{The ablation study of pre-alignment and asymmetric cross-feature fusion subnetworks on Prabhakar's datasets}
\resizebox{\linewidth}{!}{
\begin{tabular}{c|cc|cc|cccc}
\hline
\rowcolor[HTML]{EFEFEF} 
\multicolumn{1}{c|}{\cellcolor[HTML]{EFEFEF}} & \multicolumn{2}{c|}{\cellcolor[HTML]{EFEFEF}pre align} & \multicolumn{2}{c|}{\cellcolor[HTML]{EFEFEF}fusion} & \cellcolor[HTML]{EFEFEF} & \cellcolor[HTML]{EFEFEF} & \cellcolor[HTML]{EFEFEF} & \cellcolor[HTML]{EFEFEF} \\  \hhline{>{\arrayrulecolor[HTML]{EFEFEF}}->{\arrayrulecolor{black}}|----} 
\rowcolor[HTML]{EFEFEF} 
\multicolumn{1}{c|}{\multirow{-2}{*}{\cellcolor[HTML]{EFEFEF}Method}} & GLA & EFSM & ACA & $f^{guide}$ & \multirow{-2}{*}{\cellcolor[HTML]{EFEFEF}PSNR-$\mu$} & \multirow{-2}{*}{\cellcolor[HTML]{EFEFEF}PSNR-L} & \multirow{-2}{*}{\cellcolor[HTML]{EFEFEF}SSIM-$\mu$} & \multirow{-2}{*}{\cellcolor[HTML]{EFEFEF}SSIM-L} \\ \hline \hline
1 & \redcross & \redcross & \greencheck & \greencheck & 41.29 & 39.47 & {0.9728} & 0.9888 \\
2 & \greencheck & \redcross & \greencheck & \greencheck & 41.43 & 39.73 & 0.9722 & 0.9890 \\
3 & \redcross & \greencheck & \greencheck & \greencheck & 41.17 & 39.33 & 0.9726 & 0.9780 \\
4 & \greencheck & \greencheck & CA & \greencheck & \secondbest{41.61} & \secondbest{39.80} & 0.9726 & 0.9892 \\
5 & \greencheck & \greencheck & \greencheck & \redcross & 41.48 & 39.79 & 0.9723 & \secondbest{0.9893} \\
6 & \greencheck & \greencheck & \redcross & \redcross & 41.13 & 39.12 & 0.9708 & 0.9887 \\
7 & \redcross & \redcross & \redcross & \redcross & 40.81 & 38.70 & 0.9703 & 0.9784 \\
8 & w/o. Gamma & \greencheck & \greencheck & \greencheck  & 41.08 & 38.97 & \secondbest{0.9729} & {0.9884} \\
9 & w/o. Origin & \greencheck & \greencheck & \greencheck & {41.19} & {39.54} & \secondbest{0.9729} & 0.9882 \\
\rowcolor[HTML]{FFEEED} 
Ours & \greencheck & \greencheck & \greencheck & \greencheck & \best{41.80} & \best{40.14} & \best{0.9731} & \best{0.9895} \\ \hline
\end{tabular}}
\label{abstudy2}
\end{table}

\noindent \textbf{Global Luminance Alignment}. We alter the composition of the input channels and configure four scenarios to investigate the impact of GLA. As shown in Table \ref{abstudy2} (comparing Method 3, 8, 9, and Ours), incorporating GLA yields a substantial improvement in SSIM. 
By linearly aligning the intensity statistics, GLA  ensures that the subsequent fusion modules rely on geometric correspondence rather than being misled by exposure discrepancies, bringing simultaneous gains in both luminance accuracy and structural fidelity.

\begin{figure}[!t] 
    \centering 
\includegraphics[width=0.5\textwidth]{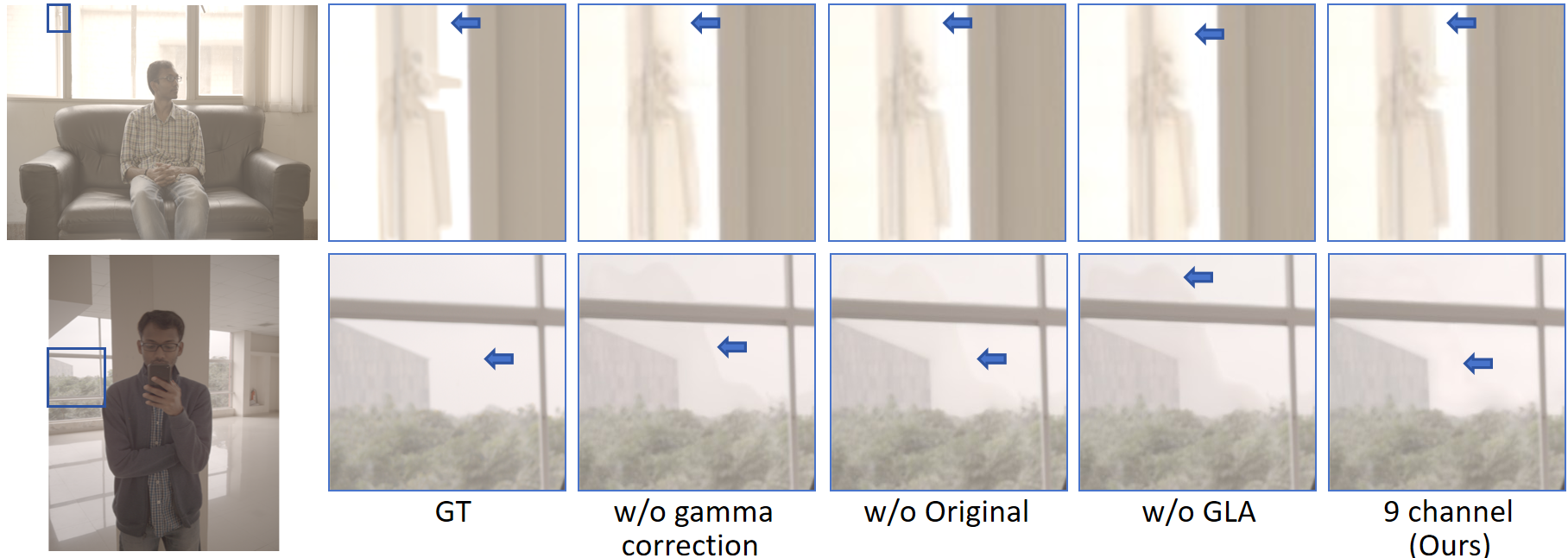} 
\caption{The contributions of global luminance alignment. 
} 
\label{visualization_abgla}
\end{figure}

The visualization results of the ablation experiments on GLA with Prabhakar's dataset are shown in Fig.~\ref{visualization_abgla}, in the first row, other variants fail to restore the overexposed details of the window frame. In the second row, ghosting occurs in the overexposed areas outside the window. The introduction of the global luminance alignment method leads to enhanced detail preservation while effectively mitigating undesirable artifacts.
Additionally, this structural alignment serves as a prerequisite for the subsequent EFSM, enabling the full model to achieve the significant performance gains observed in Table~\ref{abstudy2}.



\begin{table}[]
\caption{Experiments on changing the proportions in exposure-guided feature selection module}
 \resizebox{\linewidth}{!}{
\begin{tabular}{c|c|cccc}
\hline
\rowcolor[HTML]{EFEFEF} 
Method & Proportions & PSNR-$\mu$ & PSNR-$L$ & SSIM-$\mu$ & SSIM-$L$ \\ \hline
1 & 1:2 & 41.41 & 39.66 & \secondbest{0.9728} & 0.9791 \\
2 & 1:1 & 41.43 & 39.70 & \secondbest{0.9728} & 0.9892 \\
3 & 4:1 & 41.68 & 40.00 & 0.9725 & 0.9893 \\
4 & 8:1 & \secondbest{41.72} & \secondbest{40.02} & 0.9726 & \secondbest{0.9894} \\
5 & w/o self content & 41.37 & 39.61 & 0.9723 & 0.9769 \\
6 & w/o exposure time & 41.52 & 39.86 & 0.9723 & 0.9784  \\
\rowcolor[HTML]{FFEEED} 
Ours & 2:1 & \best{41.80} & \best{40.14} & \best{0.9731} & \best{0.9895} \\ \hline
\end{tabular}}
\label{tab2}
\end{table}

\noindent \textbf{Exposure-guided Feature Selection Module}. Table \ref{abstudy2} (comparing Methods 1, 2, 3, and Ours)  highlights a critical coupling between EFSM and GLA: \textit{1. EFSM depends on GLA for Stability.} Without GLA, EFSM fails to improve performance and even causes negative optimization. Lacking a unified luminance baseline, the module misinterprets large exposure gaps as structural noise, erroneously discarding valid features. Once GLA aligns the input statistics, it unlocks the potential of EFSM, allowing it to focus on structural selection. As shown in Fig.~\ref{abpre}, this synergy boosts PSNR-$\mu$ by 0.37 dB and significantly enhances visibility in dark regions without saturation. \textit{2. GLA depends on EFSM for Feature Discrimination.} Conversely, utilizing GLA alone leads to feature homogenization. By globally scaling pixel intensities to match the reference, GLA can mask the intrinsic SNR advantages of auxiliary frames. For instance, when high-exposure frames are linearly down-scaled to match the reference, their clean shadow details become numerically indistinguishable from the noisy shadows of the reference. EFSM serves as a critical discriminator: it explicitly identifies and preserves these high-quality regions based on their structural richness before the network conflates them with the reference features, ensuring that the alignment process does not compromise texture recovery.

Beyond the module dependency, we observe that the efficacy of feature selection is sensitive to the ratio between feature-based and exposure-based modulation coefficients. To investigate this, we conduct an in-depth analysis on six variants (Table \ref{tab2}).
\textit{1) Necessity of Explicit Exposure:} 
Explicit exposure guidance is verified to be essential. 
As evidenced in Row 6, removing the exposure input results in a performance drop of 0.28 dB compared to our method. 
This quantitative gap suggests that the network struggles to accurately infer the exposure status implicitly from image data alone, likely stemming from the physical ambiguity between object reflectance and illumination.
\textit{2) Rationale for the 2:1 Ratio:} 
Among the varied settings, the 2:1 ratio achieves the highest PSNR-$\mu$. 
Crucially, comparisons with Rows 1 and 2 reveal that over-weighting exposure degrades performance, yielding results even lower than the version without exposure guidance. 
This indicates that exposure information must serve as a regulated condition rather than a dominant feature; the 2:1 ratio represents the optimal equilibrium that resolves ambiguity without interfering with feature representation.



\begin{figure}[t] 
    \centering 
\includegraphics[width=0.47\textwidth]{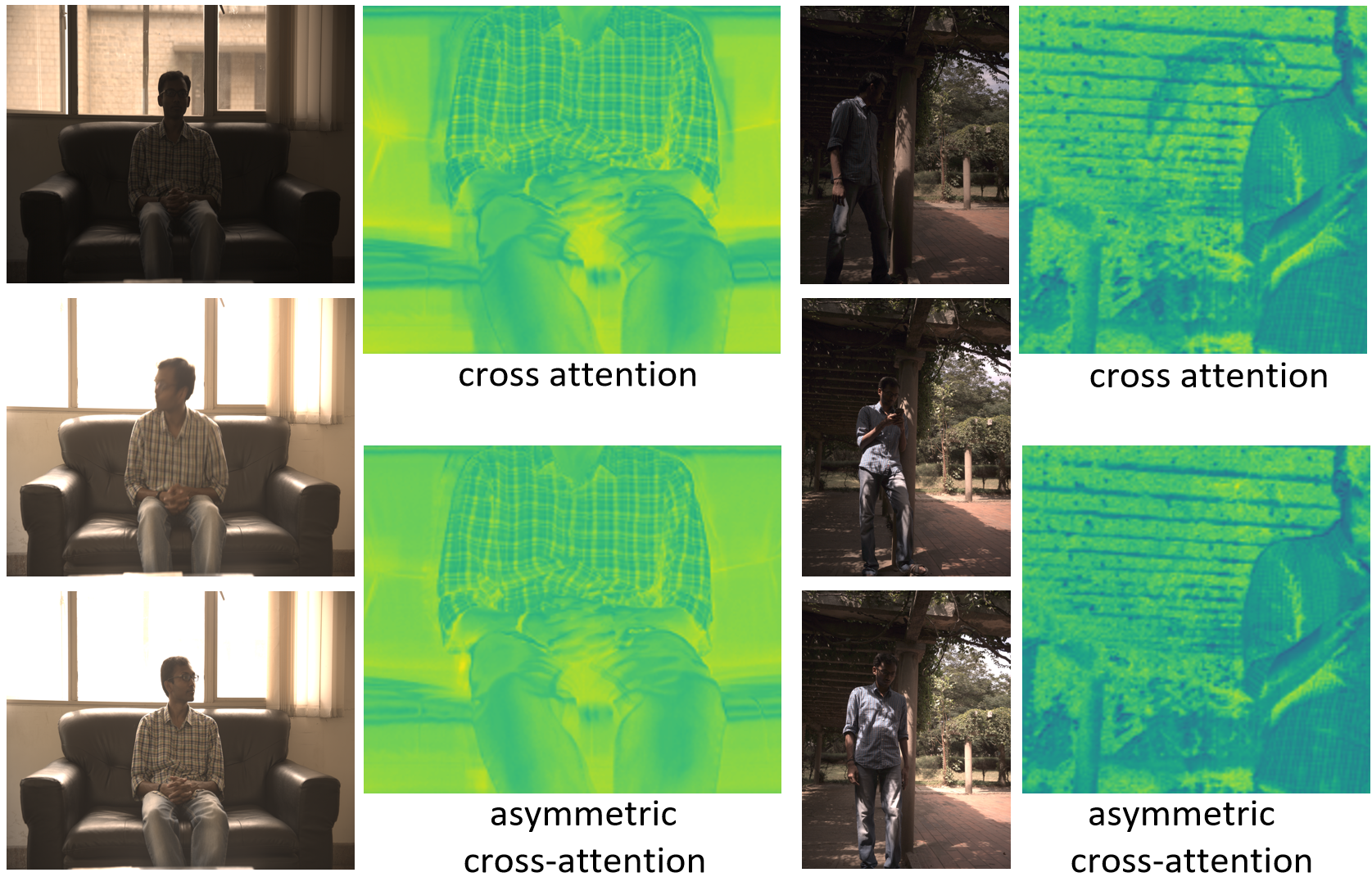} 
\caption{Visual comparison of asymmetric cross-attention and cross-attention. 
} 
\label{visualization2}
\end{figure}

\begin{figure}[!t] 
\centering 
\includegraphics[width=0.47\textwidth]{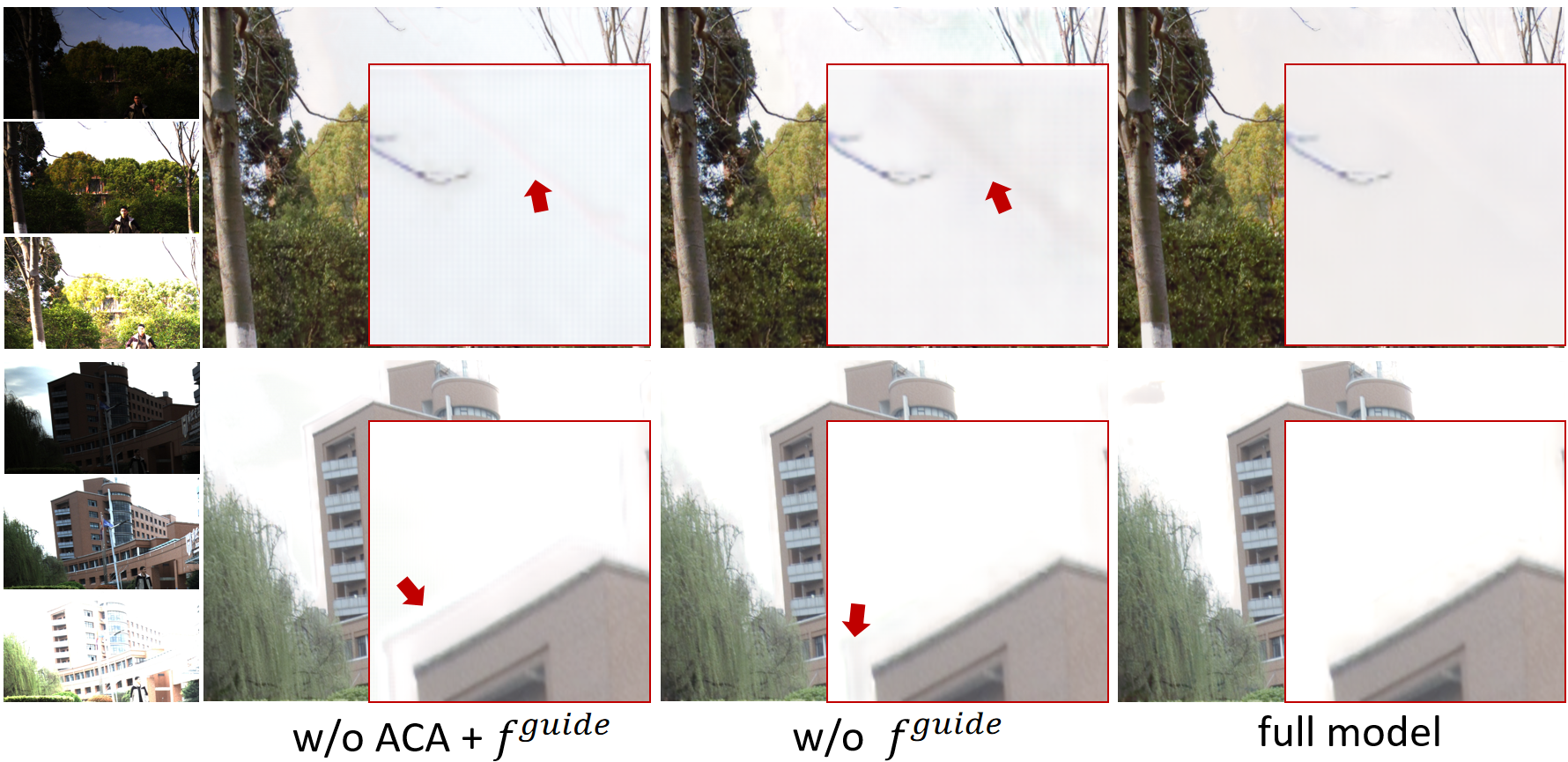} 
\caption{Visual comparison of ablation on self-captured videos. Our method reduces most ghosting artifacts.} 
\label{abvideoaca} 
\end{figure}

\noindent \textbf{Asymmetric Cross-feature Fusion Subnetwork}. Table \ref{abstudy2} demonstrates that replacing our fusion subnetwork with a generic attention block \cite{yan2019attention} causes a 0.67 dB performance drop. Qualitative results in Fig.~\ref{abvideoaca} and Fig.~\ref{visualization2} confirm our method's superior ghosting removal. Moreover, replacing standard Cross-Attention (CA) with our Asymmetric Cross-Attention (ACA) yields a 0.19 dB gain (line 5). This improvement stems from our reference-dominant design, which introduces a reference-based structural prior to suppress misaligned outliers. Additionally, the cross-scale guidance features $f^{guide}$ further refine the fusion, reducing artifacts and boosting PSNR-$\mu$ by 0.32 dB. 


\begin{table*}[!thb]
\caption{Performance Benchmarking of the Proposed Framework Across Different Hardware Platforms and Configurations.}
\resizebox{\linewidth}{!}{
\begin{tabular}{cccccccccccccc}
\hline
\textbf{Platform}                                                                       & \textbf{Resolution} & \textbf{Precision} & \textbf{Framework} & \textbf{\begin{tabular}[c]{@{}c@{}}Readout*\\ ms\end{tabular}} & \textbf{\begin{tabular}[c]{@{}c@{}}Rectification\\ (ms)\end{tabular}} & \textbf{\begin{tabular}[c]{@{}c@{}}Inference\\ (ms)\end{tabular}} & \textbf{\begin{tabular}[c]{@{}c@{}}Tone Map\\ (ms)\end{tabular}} & \textbf{\begin{tabular}[c]{@{}c@{}}Sys Memory\\ (MB)\end{tabular}} & \textbf{FPS} & \textbf{\begin{tabular}[c]{@{}c@{}}Peak VRAM\\ (MB)\end{tabular}} & \textbf{\begin{tabular}[c]{@{}c@{}}Energy\\ (J/Frame)\end{tabular}} & \textbf{Power} & \textbf{Performance} \\ \hline
\multirow{6}{*}{\textbf{\begin{tabular}[c]{@{}c@{}}Desktop \\ (RTX 4090)\end{tabular}}} & 1080p               & FP32               & PyTorch            & 8                                                              & 4                                                                     & 437                                                               & 0.006                                                            & 3902.36                                                            & 2.03         & 13797.09                                                          & 178.50                                                            & 363.21          & 44.41                \\
                                                                                        & 1080p               & FP16               & PyTorch            & 8                                                              & 4                                                                     & 251                                                               & 0.006                                                            & 3902.70                                                            & 3.73         & 6962.85                                                           & 96.50                                                             & 359.83          & 44.40                \\
                                                                                        & 1080p               & FP32               & ONNX               & 8                                                              & 4                                                                     & 434                                                               & 0.006                                                            & 3222.80                                                            & 2.30         & 22896.01                                                          & 167.42                                                            & 373.90          & 44.41                \\
                                                                                        & 1080p(tiling)       & FP32        & PyTorch            & 8                                                              & 4                                                                     & 400& 0.006                                                            & 3918.75                                                            & 2.5& 3470.36                                                           & -                                                                   & -              & 44.30                \\
                                                                                        & 1080p(tiling)       & FP16               & PyTorch            & 8                                                              & 4                                                                     & 210& 0.006                                                            & 3920.32                                                            & 4.76& 1702.42& -                                                                   & -              & 44.30                \\
                                                                                        & 4K (tiling)         & FP32               & PyTorch            & 8                                                              & 4                                                                     & 383                                                               & 0.006                                                            & 3918.80                                                            & 2.61         & 3366.09                                                           & -                                                                   & -              & 44.30                \\ \hline
\textbf{\begin{tabular}[c]{@{}c@{}}Laptop \\ (iGPU)\end{tabular}}                       & 1080p(tiling)       & FP32               & ONNX               & -                                                              & -                                                                     & 614                                                               & -                                                                & 1483.72                                                            & 1.63         & -                                                                 & -                                                                   & -              & 44.30                \\ \hline
\textbf{\begin{tabular}[c]{@{}c@{}}Mobile\\  (Jetson AGX Xavier)\end{tabular}}          & 1080p(tiling)       & FP32               & ONNX               & -                                                              & -                                                                     & 1852                                                              & -                                                                & 1809.59                                                            & 0.54         & -                                                                 & -                                                                   & -              & 44.30                \\ \hline
\end{tabular}
}
\label{speed}
\end{table*}

\subsection{limitation}
\label{limitation}

\begin{figure}[!thb] 
\centering     \includegraphics[width=0.49\textwidth]{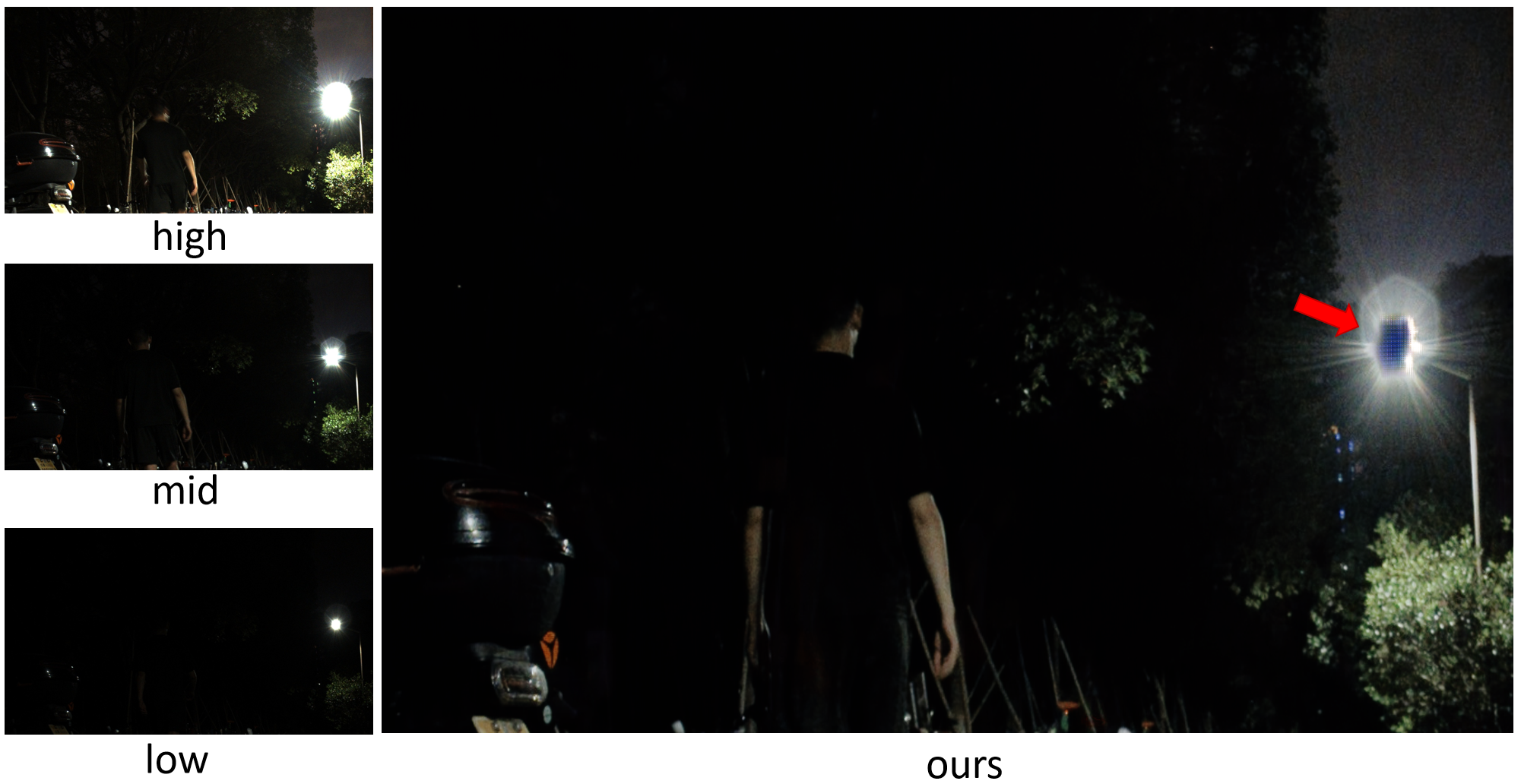} 
\caption{Visualization of a failure case in an extreme contrast nighttime scene.}
\label{figfc} 
\end{figure}

\noindent \textbf{Efficiency Analysis}.

While our asynchronous dual-camera design resolves the acquisition bottleneck for real-time capture, the reconstruction speed presents a computational trade-off. As detailed in Table~\ref{speed} and Table~\ref{videotable}, the use of Transformer-based modules for long-range dependency modeling incurs higher latency compared to lightweight CNNs.
On an RTX 4090, our method achieves $\approx$4.76 FPS at 1080p using FP16 optimization.
However, deployment on edge devices remains challenging; for instance, inference drops to 0.54 FPS on the Jetson AGX Xavier.
Future work will focus on model compression (e.g., quantization, pruning) to bridge this gap for mobile applications.

\noindent \textbf{Failure Case}. One limitation of our approach is observed in nighttime scenes with extreme contrast. As shown in Fig.~\ref{figfc}, the scene contains localized high-brightness regions (e.g., street lamps) against a nearly black background. In the reference low-exposure frame, the majority of the pixels are close to zero, providing insufficient textural cues for feature alignment, while the light sources are severely over-exposed. As a result, the network fails to properly synthesize the HDR content in these saturated areas, resulting in artifacts.



\section{CONCLUSION}
\label{conclusion}

In this work, we revisit the fundamental cause of temporal instability in alternating-exposure (AE) HDR video, which lies in the entanglement of temporal luminance anchoring with exposure-dependent detail selection, and propose a dual-stream paradigm that explicitly decouples these two roles. Specifically, we introduce an exposure-asymmetric dual-camera system (DCS) that fixes the reference exposure to stabilize frame-to-frame luminance while allowing the auxiliary stream to vary exposure for capturing high dynamic range details. To support this system, we design EAFNet to effectively suppress ghosting artifacts while preserving fine details in real-world scenes.
Extensive experiments on multiple datasets and real-world sequences demonstrate that the proposed paradigm improves both temporal stability and reconstruction quality compared with AE-based baselines and HDR image deghosting methods, while remaining cost-efficient and deployment-friendly.
Our dual-stream solution provides a promising direction for real-time HDR video capture.

\bibliography{ref2}

\end{document}